\documentclass[10pt,twocolumn,letterpaper]{article}

\usepackage{iccv}
\usepackage[dvipdfmx]{graphicx}
\usepackage{times}
\usepackage{epsfig}
\usepackage{amsmath}
\usepackage{amssymb}

\usepackage[pagebackref=true,breaklinks=true,letterpaper=true,colorlinks,bookmarks=false]{hyperref}

\iccvfinalcopy 
\ificcvfinal\fi

\begin{document}

\title{Neural Articulated Radiance Field}

\author{
    Atsuhiro Noguchi${}^\text{1}\;$   Xiao Sun${}^\text{2}\;$   Stephen Lin${}^\text{2}\;$   Tatsuya Harada${}^\text{1,3}$\\\\
    ${}^\text{1}$The University of Tokyo   $\;{}^\text{2}$Microsoft Research Asia   $\;{}^\text{3}$RIKEN
}

\maketitle

\begin{abstract}
We present Neural Articulated Radiance Field (NARF), a novel deformable 3D representation for articulated objects learned from images. While recent advances in 3D implicit representation have made it possible to learn models of complex objects, learning pose-controllable representations of articulated objects remains a challenge, as current methods require 3D shape supervision and are unable to render appearance. In formulating an implicit representation of 3D articulated objects, our method considers only the rigid transformation of the most relevant object part in solving for the radiance field at each 3D location. In this way, the proposed method represents pose-dependent changes without significantly increasing the computational complexity. NARF is fully differentiable and can be trained from images with pose annotations. Moreover, through the use of an autoencoder, it can learn appearance variations over multiple instances of an object class. Experiments show that the proposed method is efficient and can generalize well to novel poses. The code is available for research purposes at \textcolor[rgb]{1,0,1}{https://github.com/nogu-atsu/NARF}.
\end{abstract}

\begin{figure}
\begin{center}
\includegraphics[width=1.0\linewidth]{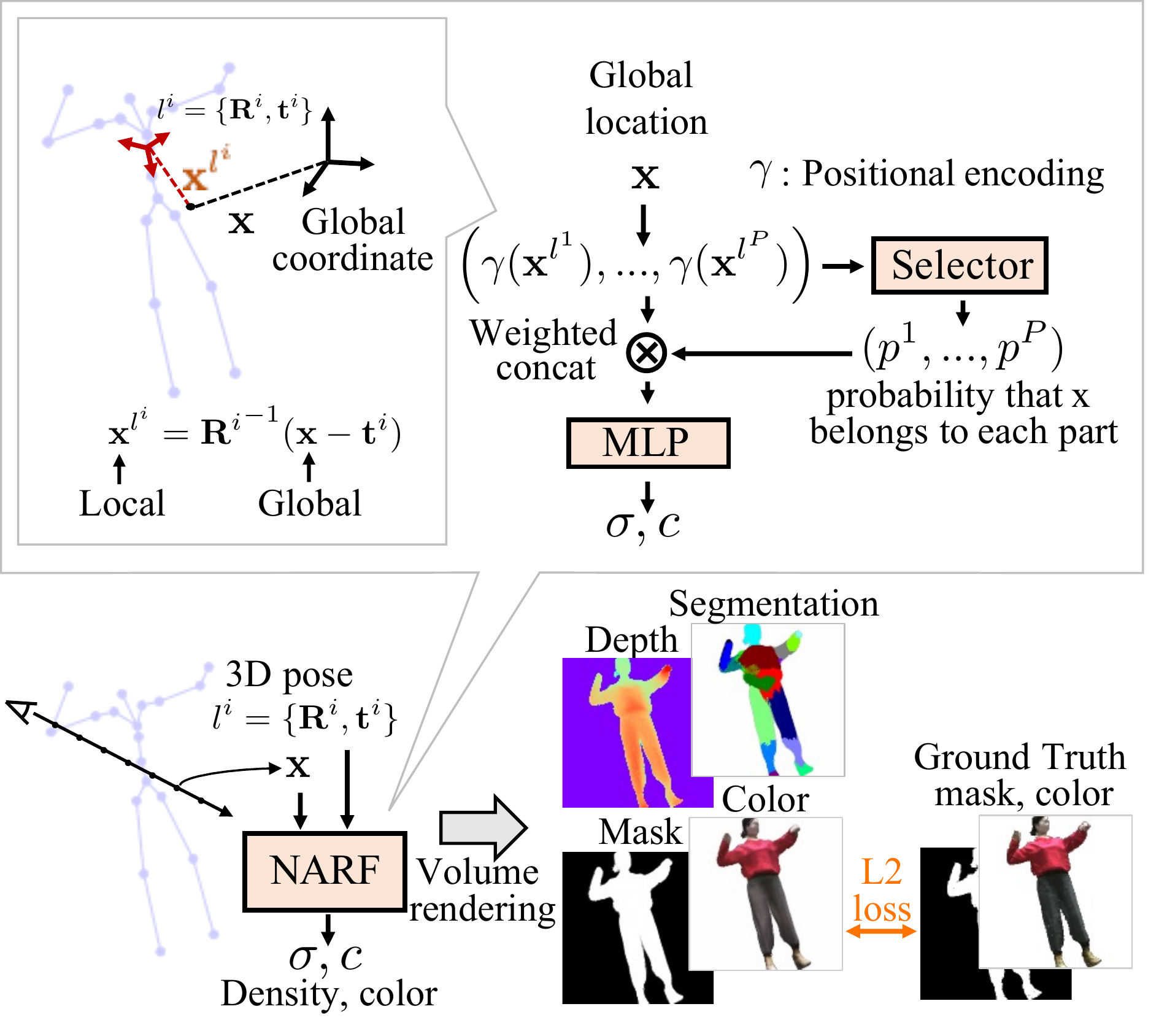}
\end{center}
\vspace{-6mm}
\caption{Training pipeline of \textbf{Disentangled NARF} ($\text{NARF}_D$). NARF is an efficient pose-aware 3D representation trained from only pose-annotated images. The learned representation is part-based and able to render novel poses of articulated 3D objects by changing the input object pose configurations.}
\label{fig:overview}
\vspace{-4mm}
\end{figure}

\section{Introduction}
In this work, we aim to learn a representation for rendering novel views and poses of 3D articulated objects, such as human bodies, from images. Our approach follows the inverse graphics paradigm~\cite{kulkarni2014inverse,kulkarni2015picture,mansinghka2013approximate,kulkarni2015deep} of analyzing an image by attempting to synthesize it with compact graphics codes. These codes are typically disentangled to allow for rendering of scenes/objects with fine-grained control over individual appearance properties such as object location, pose, lighting, texture, and shape. For the case of humans, synthesis of novel views and poses can be useful for applications such as movie making, photo editing, virtual clothing~\cite{zanfir2018human,lassner2017generative} and motion transfer~\cite{liu2019neural,chan2019everybody}.

Various inverse graphics based approaches have been
specifically designed for static scenes~\cite{wu2017neural,johnson2018image,srn}, rigid objects~\cite{dosovitskiy2015learning,zhou2016view,yan2016perspective,tatarchenko2016multi,byravan2017se3}, blend shapes for keypoints~\cite{wu2016single,chan2019everybody} and dense meshes~\cite{zanfir2018human,liu2019neural}. However, efficient deformation modeling of articulated 3D objects using neural networks remains a challenging task due to the large variance of joint locations (especially for endpoints such as hands), severe self-occlusions, and high non-linearity in forward kinematic transformations~\cite{zhou2016deep}. Though work has been done to enable explicit control over the underlying human pose~\cite{liu2019neural} and key point locations~\cite{chan2019everybody}, their neural rendering methods are either limited to 2D~\cite{chan2019everybody}, which prevents modeling of view-dependent appearance~\cite{debevec1996modeling}, or based on mesh representations~\cite{liu2019neural}, where rendering quality can be affected by the resolution of the discrete template mesh.

Recent progress on the implicit representation of 3D objects and scenes, such as signed distance functions~\cite{jiang2020local,park2019deepsdf} and occupancy fields~\cite{genova2020local,occupancy}, has greatly promoted the development of the inverse graphics paradigm. Such representations are lightweight in model size, continuous, and differentiable, making them highly practical in comparison with the previously-dominant volumetric representations~\cite{flynn2019deepview,henzler2018single,kar2017learning,mildenhall2019local,penner2017soft,srinivasan2019pushing,tulsiani2017multi,zhou2018stereo}. 
Particularly, Mildenhall et al.~\cite{nerf} propose the neural radiance field (NeRF) that takes a single continuous 5D coordinate (3D spatial location and 2D viewing direction) as input and outputs the volume density and view-dependent emitted radiance at each spatial location. Combined with a classical differentiable volume rendering technique~\cite{kajiya1984ray}, it is able to synthesize novel views by learning from a sparse set of input views of static scenes. NeRF completely discards the mesh-based representation and replaces it with a radiance-based model which can effectively and efficiently encode view-dependent appearance, enabling it to reproduce scenes of complex geometry with high fidelity.

In this paper, we extend NeRF to an articulated NeRF, called a Neural Articulated Radiance Field (NARF), to represent articulated 3D objects. Accounting for 3D articulation within the NeRF framework is a challenging problem because a complex, non-linear relationship exists between a kinematic representation of 3D articulations and the resulting radiance field, making it hard to model implicitly in a neural network~\cite{zhou2016deep}. In addition, the radiance field at a given 3D location is influenced by at most a single articulated part and its parents along the kinematic tree, while the full kinematic model is provided as input. As a result, dependencies of the output to irrelevant parts may inadvertently be learned, which is known to hurt model generalization to poses unseen in training~\cite{zeng2020srnet}.

To address these issues, we propose a method that predicts the radiance field at a 3D location based on only the most relevant articulated part. This part is identified using a set of sub-networks that output the probability for each part given the 3D location and the 3D geometric configuration of the parts. The spatial configurations of parts are computed {\em explicitly} with a kinematic model, rather than modeled implicitly in the network. A NARF then predicts the density and view-dependent radiance of the 3D location conditioned on the properties of only the selected part.
An overview of the method is shown in Fig.~\ref{fig:overview}.

The presented NARF has the following properties:
\vspace{-1mm}
\begin{itemize}
 \setlength{\itemsep}{-0.5mm}
\item It learns a disentangled representation of camera viewpoint, bone parameters, and bone pose, allowing these properties to be individually controlled in rendering.
\item A dense 3D representation is learned from a sparse set of 2D images with pose annotations of the articulated object, which could potentially be obtained through external pose estimation techniques on multi-view images with known camera parameters ~\cite{Joo_2017_TPAMI}.
\item Part segmentation is learned from images with pose annotation. Additional supervision is not needed.
\item NARF can be trained for articulated objects of various shape and appearance, through the use of an autoencoder that extracts latent shape and appearance vectors which are additionally disentangled.
\end{itemize}
\vspace{-3mm}
With this approach, it becomes possible to render both novel views and poses of articulated 3D objects from pose annotated 2D images with little increase in computational complexity.

\section{Related Work}
\vspace{-2mm}
\paragraph{Articulated 3D Shape Representations}
The deformation of articulated objects is traditionally modeled by skinning techniques~\cite{skinningcourse:2014,james2005skinning,le2012smooth,lewis2000pose} in which the location of surface mesh vertices is determined from bone transformations controlled by the kinematics~\cite{bottema1990theoretical}. Effective skinning models with subtle pose-dependent and identity-dependent deformation modeling have been developed for the human body~\cite{loper2015smpl, hesse2019learning, STAR:2020} and animals~\cite{Zuffi:CVPR:2017}. However, the representation capacity of skinning based models is limited to the resolution of the discrete template mesh, and sophisticated shading techniques~\cite{rost2009opengl} are usually required for high quality image rendering. In addition, a large amount of 3D scan data and expert supervision are required to prepare a template mesh.

Recently, Deng et al.~\cite{nasa} proposed a neural network based articulated shape representation (NASA). NASA learns the neural indicator/occupancy function~\cite{chen2019learning,occupancy,deepsdf} of every point in space, conditioned on a latent pose vector that encodes a piece-wise decomposition. NASA provides a continuous and differentiable representation for 3D articulated shapes. However, ground truth occupancy is required to train the network, and NASA does not learn appearance, a critical element for rendering.
\vspace{-4mm}
\paragraph{Articulated Pose Conditioned Image Generation} The recent advance of image generation models such as variational autoencoders (VAE)~\cite{kingma2013auto} and generative adversarial networks (GANs)~\cite{gan} provides powerful tools for generating realistic-looking images. Image generation for articulated objects (typically persons) conditioned on target poses is an important direction with various applications like movie making, photo editing, virtual clothing~\cite{zanfir2018human,lassner2017generative} and motion transfer~\cite{liu2019neural,chan2019everybody}. A majority of these works~\cite{chan2019everybody,ma2017pose,ma2018disentangled,siarohin2018deformable,villegas2017learning,esser2018variational,balakrishnan2018synthesizing} generate the image of a person in a target pose by learning a GAN model from 2D keypoint maps of the target pose. The appearance information of this person is provided by explicit concatenation with an image of this person in the target pose~\cite{ma2017pose,siarohin2018deformable}, automatically encoded for a single person~\cite{chan2019everybody} or using auto-encoders~\cite{ma2018disentangled}. These works are limited to 2D, which prevents modeling of view-dependent appearance~\cite{debevec1996modeling}. Some works~\cite{zanfir2018human} leverage the underlying 3D mesh representation and transfer appearance from one mesh to another using aligned mesh triangles. The quality of a mesh based representation is bounded by the resolution of its discrete template mesh, and a 3D template mesh is required.
\vspace{-4mm}
\paragraph{Implicit 3D representation}
Our work builds on the recent success of the implicit 3D representation. This representation is memory efficient, continuous, and topology-free, and has been used for learning 3d shape~\cite{occupancy, deepsdf, siren}, 3d texture~\cite{texturefield}, static scenes~\cite{nerf, srn}, parts decomposition~\cite{sif,ldif}, articulated objects~\cite{nasa}, deformation~\cite{dnerf, li2020neural, star, nerfies}, 3d reconstruction from sparse images~\cite{pifu,pifuhd}, and image synthesis~\cite{graf, pigan}.

Early methods required ground truth 3D geometry~\cite{occupancy, deepsdf}, but in combination with differentiable rendering, they evolved to learn from 2D images. In particular, neural radiance fields (NeRF)~\cite{nerf} is capable of learning a 3D representation of complex scenes using only multi-view posed images. However, NeRF addresses static scenes and cannot handle deformable objects. Very recently, methods have been proposed to extend NeRF to learn deformations and dynamics~\cite{dnerf, li2020neural, star, nerfies}. These models have been successful in learning deformable implicit representations using posed video frames~\cite{dnerf, li2020neural, star, nerfies}. However, these models do not take into account the structure of the object, so they cannot generate images with explicit pose control.

\section{Method}

In this section, we present neural articulated radiance field, a novel implicit representation for articulated 3D objects based on NeRF. We start by briefly reviewing the basic NeRF formulation for static scenes in Sec.~\ref{sec:nerf}. In Sec.~\ref{sec:pose-conditioned}, NeRF is extended to be conditioned on pose via a kinematic model, and a straightforward baseline is derived from this. We reformulate the pose-conditioned NeRF to allow for rigid object transformations as well as global shape variations in Sec.~\ref{sec:rigidly-transformed}. In Sec.~\ref{sec:arf}, we represent articulated 3D objects as a composition of movable rigid object parts controlled by forward kinematic rules. To achieve constant model complexity with respect to the number of object parts, we propose an efficient {\em Disentangled NARF} architecture. The training strategy is then presented in Sec.~\ref{sec:training}.

\subsection{Neural Radiance Field Revisited}
\label{sec:nerf}
A neural network is used to represent a Radiance Field such that 3D location ${\bf x} = (x, y, z)$ and 2D viewing direction ${\bf d}$ is converted to density $\sigma$ and RGB color value $c$. The density $\sigma$ acts like a differential opacity controlling how much radiance is accumulated by a ray passing through ${\bf x}$ \cite{nerf}.
\begin{equation}
F_{\Theta}:(\gamma({\bf x}), \gamma({\bf d})) \rightarrow (c, \sigma),
\end{equation}
where $\gamma(p) = [(\text{sin}(2^l\pi p),\text{cos}(2^l\pi p)]_0^L$ is a positional encoding (PE) layer that maps an input scalar into a higher dimensional space to represent high-frequency detail of the scene.
$F_{\Theta}$ consists of two ReLU MLP networks.
Specifically, the volume density $\sigma$ is a function of the location $\bf x$ only, while the RGB color $c$ is a function of both location ${\bf x}$ and viewing direction ${\bf d}$.
\begin{equation}
F_{\Theta_{\sigma}}: (\gamma({\bf x})) \rightarrow (\sigma, {\bf h}),\,\,
F_{\Theta_{c}}: ({\bf h}, \gamma({\bf d})) \rightarrow (c),
\end{equation}
where ${\bf h}$ is a hidden feature vector.

Classical volume rendering~\cite{kajiya1984ray} is used to render the color ${\bf C}({\bf r})$ of a camera ray ${\bf r}(t) = {\bf o} + t{\bf d}$ with near and far bounds $t_n$ and $t_f$, and where ${\bf o}$ denotes the camera position.
\begin{align}
T(t) &= \exp (- \int_{t_n}^{t} \sigma({\bf r}(s))ds), \\
{\bf C}({\bf r}) &= \int_{t_n}^{t_f} T(t)\sigma({\bf r}(t))c({\bf r}(t), {\bf d})dt
\end{align}
$T(t)$ denotes the accumulated transmittance along the ray.
The integrals are computed by a discrete approximation over sampled points along the ray ${\bf r}$.
\begin{align}
\label{eq:volume_rendering}
T_j = \exp \left(-\sum_{k=1}^{j-1}\sigma_k \delta_k \right),
{\bf C}({\bf r}) = \sum_{j=1}^N T_j(1 - \exp(-\sigma_j \delta_j)) c_j
\end{align}
Here, $\sigma_j$ and $c_j$ are the density and color at the $j^{th}$ point on the ray ${\bf r}$, and $\delta_j$ is the distance between the $j^{th}$ and $(j+1)^{th}$ sample points.

NeRF is trained on images of a single static scene taken from multiple views with known camera parameters. The density and color of each location are trained so that the rendered image for each of the views becomes close to its ground truth. After training, high resolution images can be synthesized from any viewpoint.

\subsection{Pose-Conditioned NeRF: A Baseline}
\label{sec:pose-conditioned}

Our goal is to extend the representation capacity of NeRF from static scenes to deformable articulated objects whose configurations can be described by a kinematic model~\cite{bottema1990theoretical}. The radiance field of a 3D location is thus conditioned on the pose configuration. Once this ``pose-conditioned NeRF" is learned, novel poses can be rendered in addition to novel views, by changing the input pose configurations.

In this work, we focus on modeling articulated objects without considering the background. Therefore, for compactness, we assume that the backgrounds are pre-cleaned.

\vspace{-4mm}
\paragraph{Kinematic Model} Formally, the kinematic model~\cite{bottema1990theoretical} represents an articulated object of $P+1$ joints, including endpoints, and $P$ bones in a tree structure where one of the joints is selected as the root joint and each remaining joint is linked to its single parent joint by a bone of fixed length.

Specifically, the root joint $J_0$ is defined by a 
global transformation matrix ${\bf T}^0$.
Let $\boldsymbol \zeta_i$ be the bone length from the $i^{th}$ joint $J_i$ to its parent, $i\in\{1,...P\}$, and $\boldsymbol \theta_i$ denotes the rotation angles of the joint with respect to its parent joint. A bone, considered as a rigid object, defines a local rigid transformation between a joint and its parent. The transformation matrix ${\bf T}^i_{local}$ is computed as
\begin{equation}\label{eq.kinematic_local}
{\bf T}^i_{local} = \text{Rot}(\boldsymbol \theta_i)\text{Trans}(\boldsymbol \zeta_i),
\end{equation}
where $\text{Rot}$ and $\text{Trans}$ are the rotation and translation matrix, respectively.
The global transformation from the root joint to joint $J_i$ can thus be obtained by multiplying the transformation matrices along the bones from the root joint to the $i^{th}$ joint:
\begin{equation}\label{eq.kinematic_global}
{\bf T}^i = (\Pi_{k \in \text{Pa}(i)} {\bf T}^k_{local}){\bf T}^0
\end{equation}
where $\text{Pa}(i)$ includes the $i^{th}$ joint and all of its parent joints along the kinematic tree. The corresponding global rigid transformation $l^i = \{R^i, {\bf t}^i\}$ for the $i^{th}$ joint can then be obtained from the transformation matrix ${\bf T}^i$.

\vspace{-4mm}
\paragraph{Baseline} The most straightforward way to condition the radiance field at a 3D location ${\bf x}$ on a kinematic pose configuration $\mathcal{P}=\{{\bf T}^0, \boldsymbol \zeta, \boldsymbol \theta\}$ is to directly concatenate a vector representing $\mathcal{P}$ as the model input.
Since the forward kinematic computation is a complex non-linear function~\cite{zhou2016deep} that is hard to simulate in neural networks, we use the transformations $l^i = \{R^i, {\bf t}^i\}$ obtained by the forward kinematics as network inputs.
\begin{align}
&F_{\Theta}^{\mathcal{P}}: (\gamma({\bf x}), \gamma(\{l^i|i=1,...,P\}), \gamma({\bf d})) \rightarrow (\sigma, c)
\label{eq:baseline}
\end{align}
We refer to this naive approach as Pose-conditioned NeRF (P-NeRF). The implementation details can be found in the supplemental material. Though P-NeRF establishes dependency between the radiance field and pose, generalization with this model is difficult because of the following two reasons.
\vspace{-1mm}
\begin{itemize}
 \setlength{\itemsep}{-0.5mm}
    \item \emph{\textbf{Implicit Transformations.}} An articulated object consists of several rigid bodies, and the surface points on the object should move with the rigid transformations of the parts when the pose changes. Therefore, the movement of points can be explicitly described using rigid body transformations of each part, but such transformations may be difficult for a neural network to learn implicitly.
    \item \emph{\textbf{Part Dependency.}}  The density at a 3D location depends only on the parameters of the bone it lies on and its parents along the kinematic tree. However, all the parameters are used to estimate the radiance field of a single location in Eq.~\ref{eq:baseline}. As the training on such 3D locations is backpropagated to all the parameters, the network may learn erroneous dependencies that do not physically exist. Correct pose predictions may still be obtained for test poses seen in the training data, but model generalization to novel poses may be degraded~\cite{zeng2020srnet}.
\end{itemize}

Towards addressing the above issues, we decompose the articulated object into $P$ rigid object parts. Each part has its own local coordinate system defined by the rigid transformation $l^i = \{R^i, {\bf t}^i\}$, which is {\em explicitly} estimated using forward kinematics, rather than modeled implicitly by a neural network. Then, we show how a rigidly transformed object part can be effectively modeled in a rigidly transformed neural radiance field (RT-NeRF) in Section~\ref{sec:rigidly-transformed}. Based on RT-NeRF, we describe in Section~\ref{sec:arf} how to train a single unified NeRF that encodes multiple parts in a manner that avoids the part dependency issue.

\subsection{Rigidly Transformed Neural Radiance Field}
\label{sec:rigidly-transformed}
Given a rigid transformation $l = \{R, {\bf t}\}$ of an object, we now estimate the radiance field in the object coordinate system where the density is constant with respect to a \emph{local} 3D location.
Formally,
\begin{equation}
F_{\Theta_{\sigma}}^l: (\gamma({\bf x}^l)) \rightarrow (\sigma, {\bf h})
\end{equation}
where ${\bf x}^l = R^{-1}({\bf x} - {\bf t})$ represents the 3D location in the local object coordinate system.

We expect the model to handle certain shape variations. For example, the limb length and thickness of a child should differ significantly from those of an adult. To account for shape variation, we further condition the model on bone parameter $\boldsymbol \zeta$.
\begin{equation}
F_{\Theta_{\sigma}}^{l, \boldsymbol \zeta}: (\gamma({\bf x}^l), \gamma(\boldsymbol \zeta)) \rightarrow (\sigma, {\bf h})
\label{eq:RT-density}
\end{equation}
Meanwhile, the color $c$ at a local 3D location may change with a transformation of the object coordinate system, as this may lead to changes in the local lighting condition.
Since the RGB color $c$ at a local 3D location should further depend on rigid transformation $l$, we use a 6D vector $\mathfrak{se}(3)$ representation $\boldsymbol \xi$ of transformation $l$ as a network input.
\begin{equation}
F_{\Theta_{c}}^{l, \boldsymbol \zeta}: ({\bf h}, \gamma({\bf d}^l), \gamma(\boldsymbol \xi)) \rightarrow (c)
\label{eq:RT-color}
\end{equation}
where ${\bf d}^l = R^{-1}{\bf d}$ is the 2D view direction in the object coordinate system.

Combining Eqs.~\ref{eq:RT-density}-\ref{eq:RT-color}, the rigidly transformed neural radiance field (RT-NeRF) defined in the $l = \{R, {\bf t}\}$ space is expressed as
\begin{equation}
F_{\Theta}^{l,\boldsymbol \zeta}:(\gamma({\bf x}^l), \gamma({\bf d}^l), \gamma(\boldsymbol \xi), \gamma(\boldsymbol \zeta)) \rightarrow (c, \sigma)
\label{eq:RT-NeRF}
\end{equation}
RT-NeRF serves as the basic building block in the neural articulated radiance field, and we will next show how it is utilized to overcome the 'Implicit Transformations' and 'Part Dependency' issues.

\subsection{Neural Articulated Radiance Field}
\label{sec:arf}

\begin{figure}
\begin{center}
\includegraphics[width=1.0\linewidth]{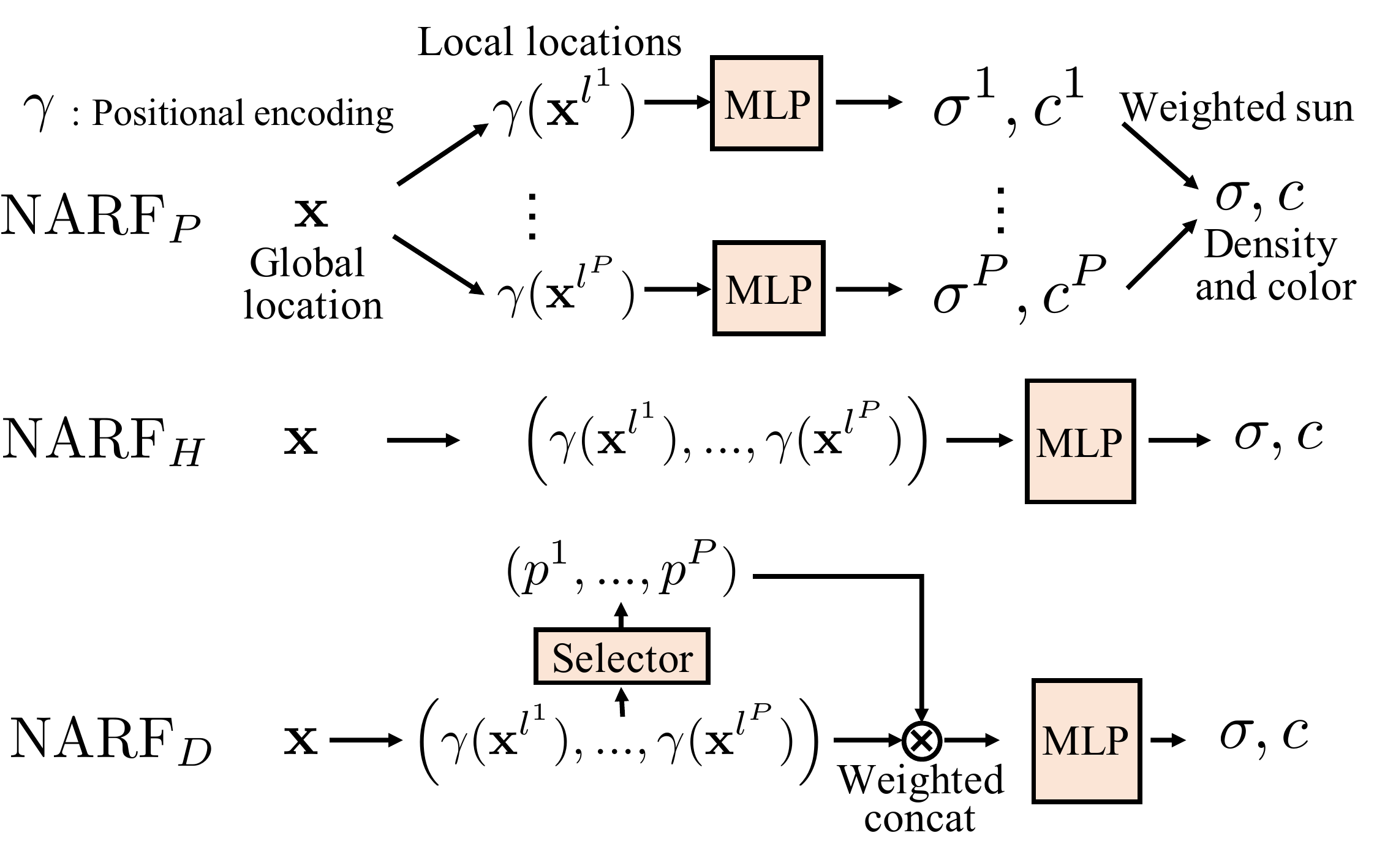}
\end{center}
\vspace{-5mm}
\caption{Three types of neural articulated radiance field. Inputs other than position ${\bf x}$ are omitted for greater clarity.}
\label{fig:arf_architecture}
\end{figure}
The proposed neural articulated radiance field (NARF) is built upon RT-NeRF. We first introduce two basic solutions, \emph{Part-Wise NARF} and \emph{Holistic NARF}, and analyze the pros and cons of each. Then, we propose our final solution named \emph{Disentangled NARF} that shares the merits of both \emph{Part-Wise} and \emph{Holistic} NARF. Conceptual figures are visualized in Fig.~\ref{fig:arf_architecture}.
\vspace{-4mm}
\paragraph{Part-Wise NARF ($\text{NARF}_P$)} Given a kinematic 3D pose configuration of an articulated object $\{{\bf T}^0,\boldsymbol \zeta,\theta\}$, we first compute the global rigid transformation $\{l^i|i=1,...,P\}$ for each rigid part using the forward kinematics in Eqs.~\ref{eq.kinematic_local}-\ref{eq.kinematic_global}. To estimate the density and color $(\sigma, c)$ of a global 3D location ${\bf x}$ from a 2D viewing direction ${\bf d}$, we train a separate RT-NeRF, $F_{\Theta^i}^{l^i,\boldsymbol \zeta}$,
\begin{align}
&{\bf x}^{l^i} = {R^i}^{-1}({\bf x} - {\bf t}^i), {\bf d}^{l^i} = {R^i}^{-1}{\bf d}\\
&F_{\Theta^i}^{l^i,\boldsymbol \zeta}:(\gamma(x^{l^i}), \gamma({\bf d}^{l^i}), \gamma(\boldsymbol \xi^i), \gamma(\boldsymbol \zeta)) \rightarrow (c^i, \sigma^i),
\end{align}
for each part using Eq.~\ref{eq:RT-NeRF} and combine the densities and colors $\{\sigma^i, c^i|i=1,...,P\}$ estimated by different RT-NeRFs into one. We denote this approach as \emph{\textbf{Part-Wise NARF}} ($\text{NARF}_P$).

Since a surface point of an object can belong to only one of the object parts, only one of the estimates in $\{\sigma^i, c^i|i=1,...,P\}$ should be nonzero. The density and color $(\sigma, c)$ of a global 3D location ${\bf x}$ can be determined by taking the estimate with the highest density. However, the max operation is not differentiable, so we instead use the softmax function, which is a differentiable weighted sum over all the estimates:
\begin{align}\label{eq.softmax_sigma}
\sigma = \frac{\sum_{i=1}^P \exp(\sigma^i/\tau) \sigma^i}{\sum_{i=1}^P \exp(\sigma^i/\tau)},
c = \frac{\sum_{i=1}^P \exp(\sigma^i/\tau) c^i}{\sum_{i=1}^P \exp(\sigma^i/\tau)},
\end{align}
where $\tau$ is the temperature parameter of the softmax function. Volume rendering is then applied using the combined density $\sigma$ and color $c$ to generate the rendered color ${\bf C}({\bf r})$ by Eq.~\ref{eq:volume_rendering}. Since the rendering and softmax operations are both differentiable, the image reconstruction loss can pass gradients to all the RT-NeRF models for effective training.

We note that, in addition to color ${\bf C}({\bf r})$, the foreground mask ${\bf M}({\bf r})$ can also be estimated as an integral of the opacity along the camera ray:
\begin{equation}\label{eq.mask_rendering}
{\bf M}({\bf r}) = \sum_{j=1}^N T_j(1 - \exp(-\sigma_j \delta_j))
\end{equation}
We can further render a segmentation image that indicates which RT-NeRF (object part) is used for rendering each pixel:
\begin{align}
    &s_j({\bf r}) = \text{arg}\max_i \{\sigma_j^i|i \in [1,P]\}\\
&{\bf S}^{i}({\bf r}) = \sum_{j=1}^N T_j(1 - \exp(-\sigma_j \delta_j)) (s_j({\bf r})==i)
\label{eq:part_segmentation}
\end{align}
where $s_j$ is the index of the part with the greatest density, and ${\bf S}^i$ denotes a segmentation mask for the $i^{th}$ part.

\textbf{Discussion.} The $\text{NARF}_P$ approach models the rigidly transformed parts of an articulated object in separate RT-NeRFs, where each part has a consistent radiance field under different 3D pose configurations. As the rigid transformation of each RT-NeRF is computed explicitly via forward kinematics, rather than implicitly within the network, the issue of implicit transformations and part dependency are avoided. 
The part dependency issue is also addressed by taking the estimate with the highest density while suppressing the contribution of other parts for a global 3D location in Eq.~\ref{eq.softmax_sigma}. However, its computation is inefficient for the following reasons.
\vspace{-1mm}
\begin{itemize}
 \setlength{\itemsep}{-0.5mm}
    \item The computational cost is proportional to the number of object parts, limiting the representation capacity for complex articulated objects.
    \item Training is dominated by the large number of zero density point samples. As a surface point on an object can belong to only one of the object parts, it will be trained as a zero density sample for the remaining parts. Since parts with small densities do not affect the value of the equation very much, it is not really necessary to calculate the density of those parts.
\end{itemize}
\vspace{-4mm}
\paragraph{Holistic NARF ($\text{NARF}_H$)}
To address the above issues, we present another approach that combines the inputs of the RT-NeRF models in $\text{NARF}_P$ then feeds them as a whole into a single NeRF model for direct regression of the final density and color $(\sigma, c)$. We call this approach \emph{\textbf{Holistic NARF}} ($\text{NARF}_H$). Formally,
\begin{align}
    F_{\Theta_{\sigma}}^{l, \boldsymbol \zeta}: \text{Cat}(\{\gamma({\bf x}^{l^i})|i \in [1, P]\}, \gamma(\boldsymbol \zeta)) \rightarrow (\sigma, {\bf h}),\\
    F_{\Theta_{c}}^{l, \boldsymbol \zeta}: \text{Cat}({\bf h}, \{(\gamma({\bf d}^{l^i}), \gamma(\boldsymbol \xi^i))|i \in [1, P]\}) \rightarrow (c),
\end{align}
where Cat denotes the concatenation operator.

\textbf{Discussion.} There is only a single NeRF model trained in $\text{NARF}_H$. The computational cost is almost constant to the number of object parts and the zero density problem is naturally avoided. However, unlike \emph{\textbf{Part-Wise NARF}}, $\text{NARF}_H$ does not satisfy \emph{\textbf{Part Dependency}}, because all parameters are considered for each 3D location. Moreover, object part segmentation masks cannot be generated from Eq.~\ref{eq:part_segmentation} without part dependencies.

\vspace{-4mm}
\paragraph{Disentangled NARF ($\text{NARF}_D$)}
We propose \emph{\textbf{Disentangled NARF}} ($\text{NARF}_D$) which shares the merits of both $\text{NARF}_P$ and $\text{NARF}_H$ while avoiding their weaknesses by introducing a selector $\mathcal{S}$.

The selector $\mathcal{S}$ identifies which object part a global 3D location ${\bf x}$ belongs to. $\mathcal{S}$ consists of $P$ lightweight sub-networks for each part. For the $i^{th}$ part, a sub-network $O_{\Gamma}^i$ takes the local 3D position of ${\bf x}$ in $l^i=\{R^i, {\bf t}^i\}$ and the bone parameter $\boldsymbol \zeta$ as input and outputs the probability $p^i$ of ${\bf x}$ belonging to the $i^{th}$ part. Since ${\bf x}$ should be assigned to only one of the object parts, the softmax activation is used to normalize the selector's outputs:
\begin{align}\label{eq:selector_occupancy}
    &O_{\Gamma}^i: (\gamma({\bf x}^{l^i}), \gamma(\boldsymbol \zeta)) \rightarrow (o^i), \,
    &p^i = \frac{\exp(o^i)}{\sum_{k=1}^P \exp(o^k)}
\end{align}
It can be seen that $O_{\Gamma}^i$ is actually an occupancy network~\cite{occupancy,nasa} defined \emph{in the local object coordinate system}. Comparing this with NASA~\cite{nasa}, NASA's occupancy networks learn absolute occupancy values to estimate an explicit surface, but our networks learn relative occupancy values to other parts for part selection. For implementation, we use a two-layer MLP with ten hidden nodes for each occupancy net, which is lightweight yet effective.

\emph{\textbf{Disentangled NARF}} is defined by (softly) masking out the irrelevant parts in the concatenated input using the outputs $p^i$ of the selector. 
\begin{align}\label{eq:narfd_sigma}
    &F_{\Theta_{\sigma}}^{l, \boldsymbol \zeta}: \text{Cat}(\{\gamma({\bf x}^{l^i}) * p^i|i \in [1, P]\}, \gamma(\boldsymbol \zeta)) \rightarrow (\sigma, {\bf h}),\\
    &F_{\Theta_{c}}^{l, \boldsymbol \zeta}: \text{Cat}({\bf h}, \{(\gamma({\bf d}^{l^i}) * p^i, \gamma(\boldsymbol \xi^i) * p^i)|i \in [1, P]\}) \rightarrow (c)
\end{align}
Note that though we have removed the dependency on irrelevant parts by masking their inputs, the resulting input is still in the form of a concatenation.
This is done purposely because all the bones share a single NeRF, which needs to distinguish the different bones in order to generate the corresponding density and color. Different bones are distinguished by dimensions of the concatenated vector that represent them, similar to part identity encoding. The detailed network architecture can be found in Fig.~\ref{fig:network_architecture} of the supplemental material.

Since the selector outputs the probabilities of a global 3D location belonging to each part, we can generate the segmentation mask by selecting the locations occupied by a specific part followed by Eq.~\ref{eq:part_segmentation}:
\begin{align}
    &s_j({\bf r}) = \text{arg}\max_i \{p_j^i|i \in [1,P]\}
\end{align}

\subsection{Training Details}
\label{sec:training}
The positional encoding dimensions we set for the 3D location ${\bf x}$ and other parameters are 10 and 4 respectively.
During training, at each optimization iteration, we randomly sample a batch of camera rays from the set of all pixels, and then follow the hierarchical volume sampling strategy of the original NeRF~\cite{nerf} to query $N$ samples for each ray. With known kinematic 3D pose configuration $\{{\bf T}^0,\boldsymbol \zeta,\theta\}$, the samples' densities and colors are estimated by the NARF model. Volume rendering is then used to render the color ${\bf C}({\bf r})$ and mask ${\bf M}({\bf r})$ of this ray using Eq.~\ref{eq:volume_rendering} and~\ref{eq.mask_rendering}, respectively. The loss is the total squared error between the rendered and true pixel colors and masks.
\begin{align}\label{eq.loss}
\mathcal L = \sum_{{\bf r} \in \mathcal{R} } [||\hat{\bf C}({\bf r}) - {\bf C}({\bf r})||_2^2 + ||\hat{\bf M}({\bf r}) - {\bf M}({\bf r})||_2^2]
\end{align}
where $\mathcal{R}$ is the set of rays in each batch, $\hat{\bf C}$ and $\hat{\bf M}$ are the ground truth color and foreground mask. In the supplemental material, we empirically show that the extra mask loss helps to learn a cleaner background. Other training details on the learning rate, batch size and optimizer can be found in the supplemental material.

\section{Results of Training on a Single Object}
In this section, we evaluate our model in the case of a single articulated 3D object.
\vspace{-4mm}
\paragraph{Dataset and Settings}
We create our own synthetic dataset of human bodies for experimentation. It consists of two persons, one male and one female, selected from the human 3D textured mesh (THUman) dataset~\cite{Zheng2019DeepHuman}. Each person has 56 and 48 different poses, 26 of which are used for training and the others for testing. We render 100 images with various orientations and scaling of each mesh for training and 20 for testing, ending up with 2600 training images for each person. Note that the viewpoint distribution of training and testing sets are the same under this setting. We denote this test data setting as the \textbf{novel pose/same view} setting. All rendered images have a resolution of 128 $\times$ 128.
Additionally, we introduce three other test settings for a more comprehensive comparison. The \textbf{same pose/same view} setting uses testing images rendered from the same poses and same viewpoint distribution as in training. The \textbf{novel pose/novel view} setting uses novel poses and a different viewpoint distribution than in training. Finally, the \textbf{same pose/novel view} setting uses the same poses but the viewpoint distribution is different from training.
The kinematic 3D pose configurations are inferred from the SMPL model parameters provided by the THUman dataset and we use bone length as the bone parameter $\zeta$ in Eq.~\ref{eq:RT-density}.

\begin{table*}[t]
\begin{center}
\scalebox{0.70}{
  \begin{tabular}{l|ccc|rrr|rrr|rrr|rrr} \hline
  & \multicolumn{3}{c|}{Cost} & \multicolumn{3}{c|}{Same pose, same view} & \multicolumn{3}{c|}{Novel pose, same view} & \multicolumn{3}{c|}{Same pose, novel view} & \multicolumn{3}{c}{Novel pose, novel view} \\ \hline
   Method & \#Params & \#FLOPS & \#Memory &Mask$\downarrow$ & PSNR $\uparrow$ & SSIM $\uparrow$ &  Mask $\downarrow$ & PSNR $\uparrow$ & SSIM $\uparrow$ & Mask $\downarrow$ & PSNR $\uparrow$ & SSIM $\uparrow$ & Mask $\downarrow$ & PSNR $\uparrow$ & SSIM $\uparrow$ \\ \hline \hline

CNN                           &15.6M& -   &-                & 76.9&29.12&0.9429&134.8&27.30&0.9211&\underline{365.9}&\underline{25.19}&\underline{0.8532}&\underline{392.2}&\underline{24.53}&\underline{0.8470}\\
P-NeRF                        &0.85M&156M &356K             & 778.7&21.42&0.8006&1077.0&20.42&0.7696&844.9&21.19&0.7897&1110.1&20.27&0.7648\\
D-N{\footnotesize A}RF        &0.66M&121M &382K             & 2182.6&18.90&0.1143&2308.2&18.81&0.1140&2137.3&19.09&0.1144&2241.3&18.88&0.1133\\ \hline
$\text{NARF}_P$   &11.8M&\underline{2140M}&\underline{6544K}& 92.0&28.56&0.9258&116.2&26.83&0.9052&101.5&27.54&0.9144&125.8&26.50&0.9104\\
$\text{NARF}_H$               &1.06M&197M &344K             & 55.6&29.91&0.9470&\underline{376.8}&\underline{24.09}&\underline{0.8665}&70.5&28.81&0.9370&\underline{374.6}&\underline{23.98}&\underline{0.8646}\\
$\text{NARF}_D$               &1.10M&205M &382K             & {\bf 50.5}&{\bf 30.86}&{\bf 0.9586}&{\bf 114.4}&{\bf 27.93}&{\bf 0.9317}&{\bf 64.1}&{\bf 29.44}&{\bf 0.9466}&{\bf 123.8}&{\bf 27.24}&{\bf 0.9230}\\
\hline
  \end{tabular}}
    \caption{Quantitative comparison for a single object.
    Best results in {\bf bold}.
    }
    \label{single_26_results}
\end{center}
\vspace{-8mm}
\end{table*}

\begin{figure*}
\begin{center}
\includegraphics[width=1.0\linewidth]{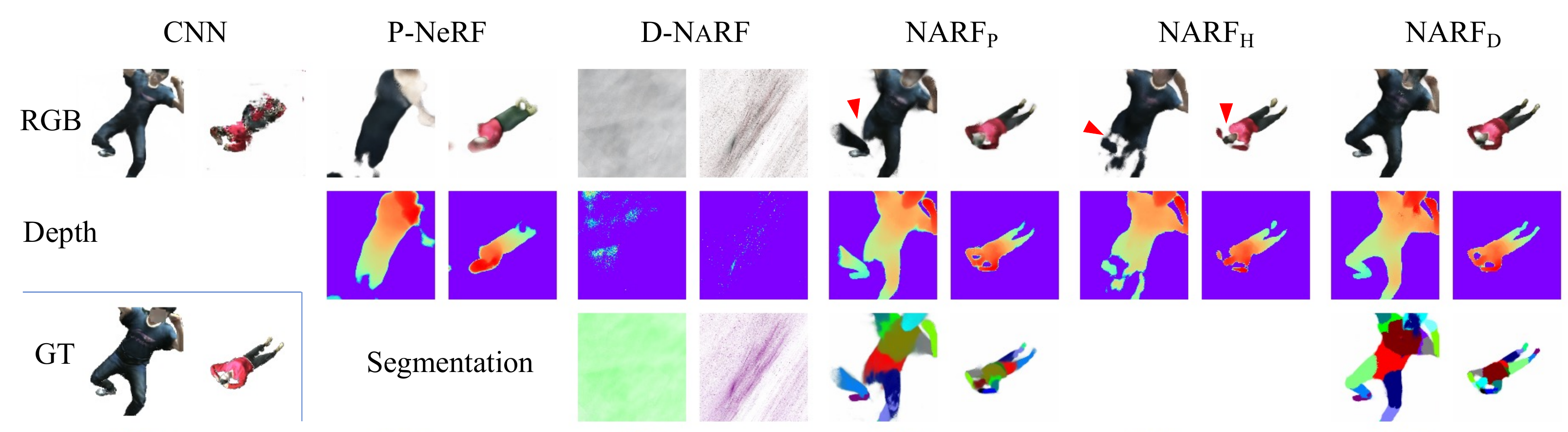}
\end{center}
\vspace{-5mm}
\caption{Generative results comparison for a single object with a novel pose and novel view. One model per person identity. Triangles point to areas that should be noted. $\text{NARF}_D$ best generalizes to novel view/novel poses among the six methods.}
\label{fig:supervised_result}
\vspace{-2mm}
\end{figure*}

\begin{figure}
\begin{center}
\includegraphics[width=1.0\linewidth]{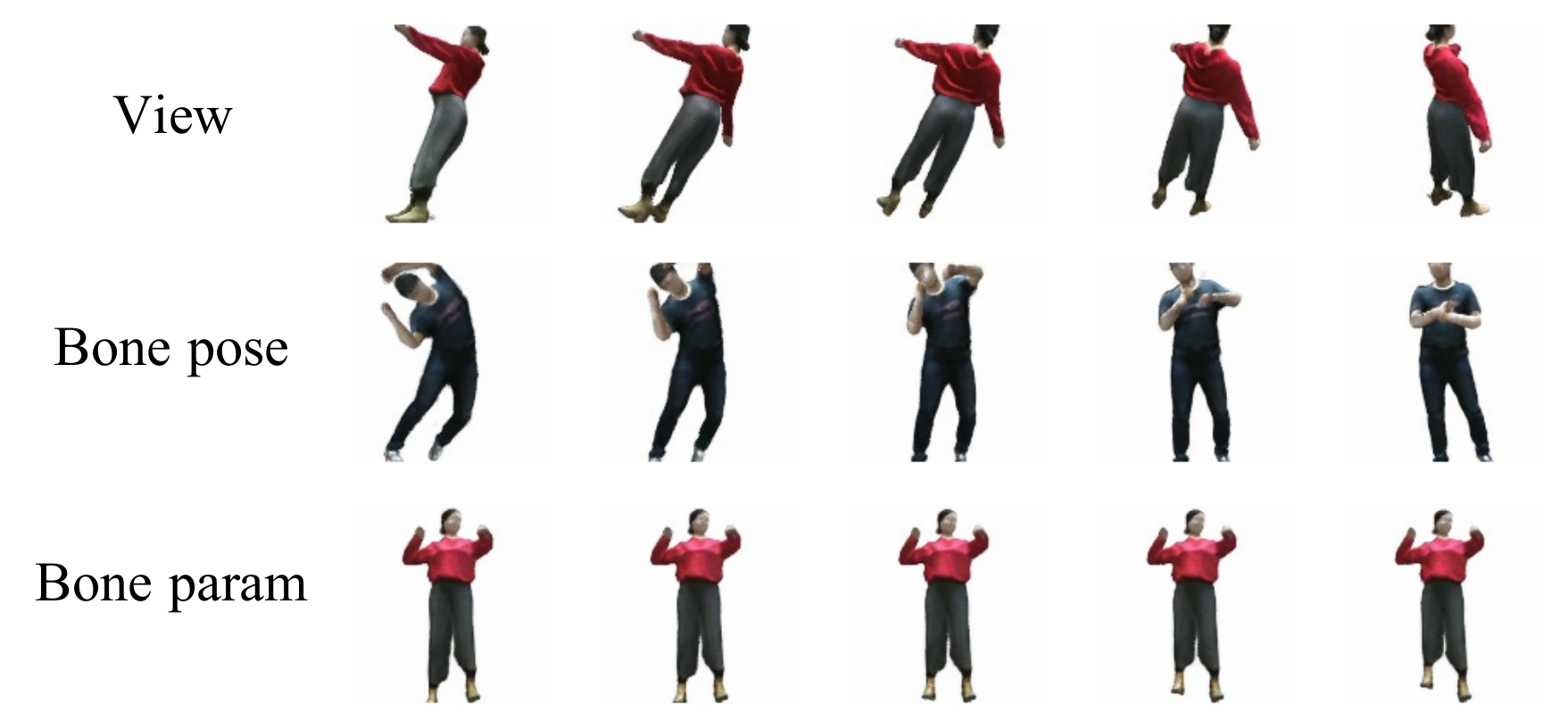}
\end{center}
\vspace{-5mm}
\caption{Disentangled representations learned by $\text{NARF}_D$. For bone parameter interpolation, the right leg length is interpolated.}
\label{fig:supervised_disentangle_result}
\vspace{-2mm}
\end{figure}

\vspace{-4mm}
\paragraph{Metrics} Three metrics are used to evaluate performance. The peak signal to noise ratio (PSNR) and structural similarity index (SSIM)~\cite{ssim} are two commonly used evaluation metrics for image reconstruction (higher is better). In addition, we introduce the L2 distance error of mask images (Mask), which better describes how close the 3D shape of the rendered object is to the ground truth (lower is better).

\vspace{-4mm}
\paragraph{Baselines}
In addition to the three variants of NARF ($\text{NARF}_P$, $\text{NARF}_H$ and $\text{NARF}_D$),
three other baselines are included for comparison. The first is a 2D \emph{\textbf{CNN-based}} method similar to \cite{chan2019everybody} that generates the target subject image from ``pose stick figures". The pose stick figures in our case are generated by projecting the 3D joints into the 2D image (with given camera parameters) then adding lines to connect these 2D keypoints. The second is the \emph{\textbf{P-NeRF}} method described in Sec.~\ref{sec:pose-conditioned}. The third one is \emph{\textbf{D-N{\footnotesize A}RF}}, a simple extension of D-N{\footnotesize E}RF~\cite{nerfies} to articulated objects. D-N{\footnotesize A}RF aims to learn the mapping $\Psi:{\bf x} \rightarrow {\bf x}'$ that transforms a given point to its position in a canonical shape space. In our implementation, a static NeRF model for a canonical pose $\mathcal{P}^c$ is learned, then a mapping network~\cite{nerfies} $\Psi$ estimates the deformation field between the scene of a specific pose instance $\mathcal{P}$ and the scene of the canonical pose $\mathcal{P}^c$. The details of the three baselines can be found in the supplemental material.

\vspace{-4mm}
\paragraph{Results}
Quantitative comparison results are given in Table \ref{single_26_results}. \#Params, \#FLOPS, and \#Memory denote the number of parameters, floating point operations per ray, and number of elements to preserve during forward propagation per ray, which is proportional to the memory cost.

It can be seen that our method, \textbf{$\text{NARF}_D$}, outperforms the others under all the evaluation metrics and test data settings (best results shown in \textbf{bold}). Particularly, it exhibits high performance under novel pose and/or novel view settings (slight performance drop on SSIM within 4\%) with low computational cost (close to a single NeRF model, P-NeRF). Hence, we can conclude that $\text{NARF}_D$ effectively and efficiently learns the radiance field of an articulated 3D object and the model generalizes to novel poses and views with high fidelity.

In contrast, all the other methods are deficient in one way or another. The \textbf{CNN-based} method fails when tested under novel views (10\% performance drop on SSIM) since it is difficult to learn an effective 3D representation from 2D inputs. \textbf{P-NeRF} and \textbf{D-N{\footnotesize A}RF} fail in almost all cases mainly due to both the Implicit Transformations and Part Dependency issues.
\textbf{$\text{NARF}_P$} exhibits good performance and generalization ability but requires much more computation ($10 \times$ \#FLOPS and $17 \times$ \#Memory of \textbf{$\text{NARF}_D$}). \textbf{$\text{NARF}_H$} is less effective when tested on novel poses (8\% performance drop on SSIM) due to the Part Dependency issue.

Qualitative results under the \textbf{novel pose/novel view} setting are shown in Fig.~\ref{fig:supervised_result}.
Rendered RGB images (first row), depth maps (second row), and part segmentation maps (third row), as well as ground truth RGB images (bottom left corner) are displayed. It can be seen that the NeRF based methods (except for the CNN-based one) can obtain depth images and the ``part dependent'' methods ($\text{NARF}_P$ and $\text{NARF}_D$) can obtain segmentation maps. Our final solution $\text{NARF}_D$ generates higher quality RGB, depth and segmentation maps for novel views and poses than the others. Moreover, as shown in Fig.~\ref{fig:supervised_disentangle_result}, $\text{NARF}_D$ learns a disentangled representation of camera viewpoint, bone parameters and pose, allowing these appearance properties to be individually controlled in rendering.

\section{Appearance Variation with Autoencoder}
\label{sec:autoencoder}

In this section, we train an autoencoder based on NARF to model shape and appearance variation among multiple articulated objects.
The autoencoder consists of an encoder and decoder.
First, a 2D CNN-based encoder is used to generate a latent vector ${\bf z}$ from an input image. The obtained latent vector together with the given camera viewpoint and human pose are fed into our NARF based decoder to reconstruct the input image.

Following the implementation for the NeRF-based generator~\cite{graf}, we first decompose ${\bf z}$ into a shape latent vector ${\bf z}_s$ and an appearance latent vector ${\bf z}_a$. Then, ${\bf z}_s$ is concatenated to the density-dependent inputs, namely, the positionally encoded location ${\bf x}$ and bone parameters $\boldsymbol \zeta$. Meanwhile, ${\bf z}_a$ is concatenated to the color-dependent inputs, namely, the positionally encoded view direction ${\bf d}$ and the local transformation $\boldsymbol \xi$.
Specifically, when combining the autoencoder with the $\text{NARF}_D$ model, we have
\begin{equation}
    F_{\Theta_{\sigma}}^{l, \boldsymbol \zeta, {\bf z}}: \text{Cat}(\{\gamma({\bf x}^{l^i}) * p^i|i \in [1, P]\}, \gamma(\boldsymbol \zeta), {\bf z}_s) \rightarrow (\sigma, {\bf h}),
\end{equation}
\vspace{-5mm}
\begin{equation}
    F_{\Theta_{c}}^{l, \boldsymbol \zeta, {\bf z}}: \text{Cat}({\bf h}, \{(\gamma({\bf d}^{l^i}) * p^i, \gamma(\boldsymbol \xi^i) * p^i)|i \in [1, P]\}, {\bf z}_a) \rightarrow (c)
\end{equation}
The encoder and decoder are trained jointly using the same loss in Eq.~\ref{eq.loss}. For the experiment, we use the best performing $\text{NARF}_D$ by default. Comparison results for other models can be found in the supplemental material.

\vspace{-4mm}
\paragraph{Dataset}
We create another synthetic human body dataset from THUman~\cite{Zheng2019DeepHuman} for experimentation. All of the males (112 in total) and poses (35 on average for each person) in the THUman dataset are used. We render 10 images for each pose with randomly sampled viewpoints to generate 35450 images for training and 3940 images for evaluation. All rendered images have a resolution of 128 $\times$ 128.

\begin{figure}
\begin{center}
\includegraphics[width=1.0\linewidth]{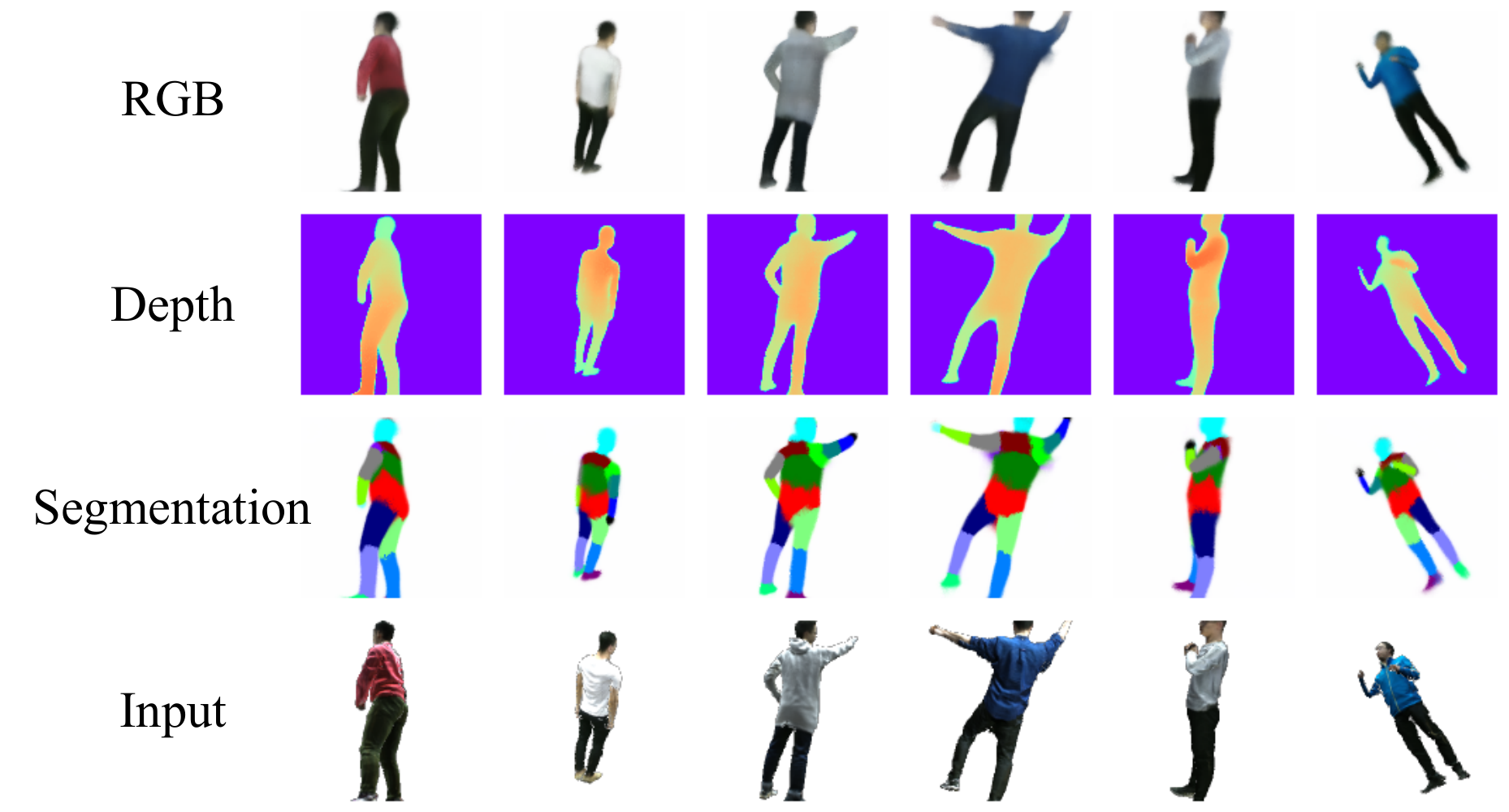}
\end{center}
\vspace{-5mm}
\caption{Single-view reconstruction results from $\text{NARF}_D$ with an autoencoder. All outputs are generated from a single model.}
\label{fig:autoencoder_result}
\vspace{-2mm}
\end{figure}

\begin{figure}
\begin{center}
\includegraphics[width=0.9\linewidth]{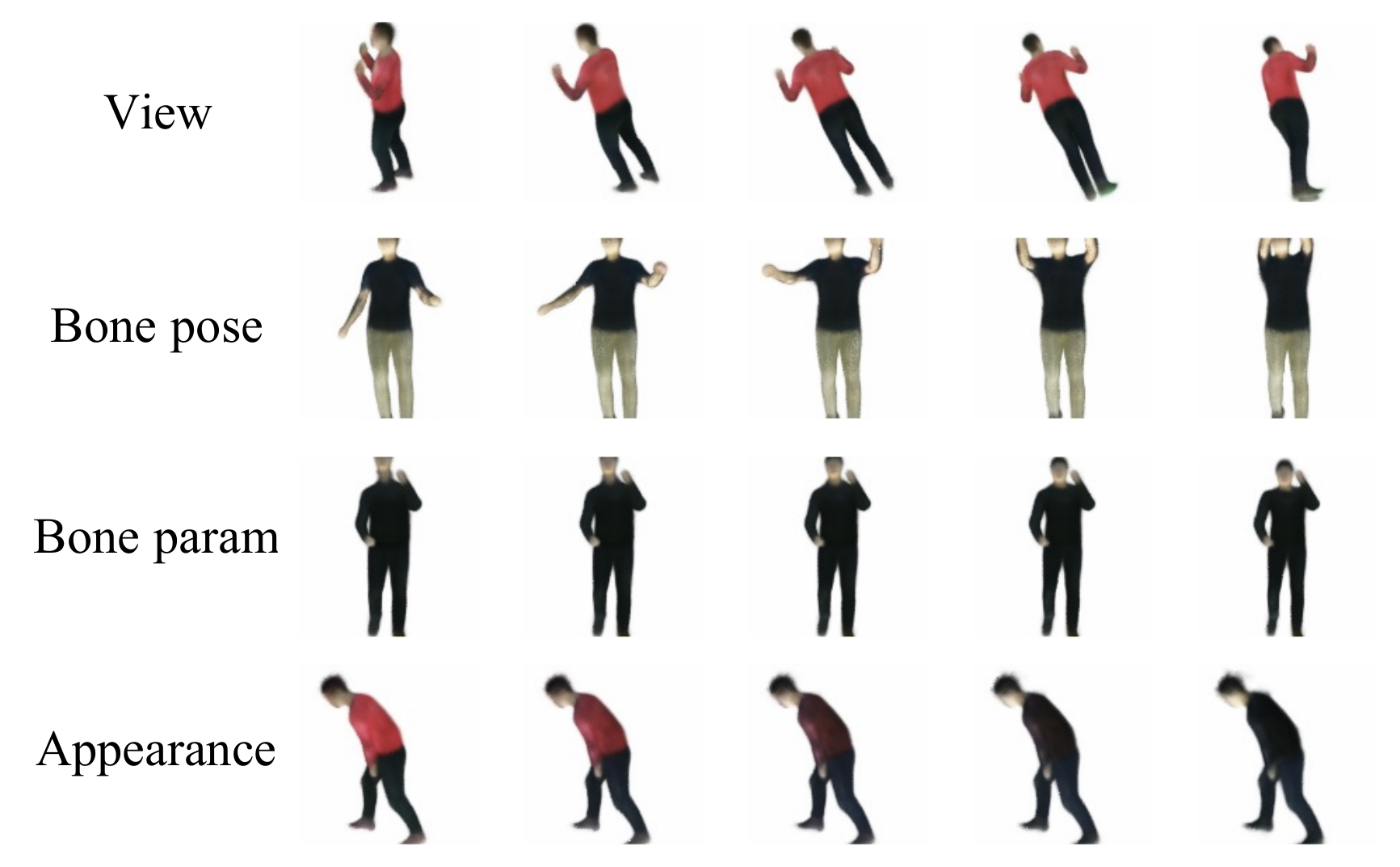}
\end{center}
\vspace{-5mm}
\caption{Disentangled representations learned with an autoencoder. For bone parameter interpolation, the head position is interpolated.}
\label{fig:ae_disentangle_result}
\vspace{-2mm}
\end{figure}

\vspace{-4mm}
\paragraph{Results}
Fig.~\ref{fig:autoencoder_result} shows the reconstructed RGB images as well as the additional depth images and segmentation maps generated from the input RGB images. A single autoencoder is trained for all objects, indicating that appearance variation is effectively modeled.
Fig.~\ref{fig:ae_disentangle_result} shows that the NARF-based autoencoder learns a disentangled representation of camera viewpoint, bone parameters, human pose, and color appearance, allowing these properties to be individually controlled in rendering. For color appearance, it is controlled by replacing the appearance latent vector ${\bf z}_a$ with that of another person.
Additional results can be found in the supplemental material.

\section{Conclusion and Future work}
In this paper, we propose a method for learning implicit representations for articulated objects. We show that it is possible to learn explicitly controllable representations of viewpoint, pose, bone parameters, and appearance from 3D pose annotated images. Although pose annotation is required to train the model, the model is differentiable and thus may be extended to reduce the required supervisory information, for example, by simultaneously training 3D pose estimation and segmentation with the model. In addition, since the proposed representation provides explicit 3D shape and part segmentation, it may be applied to unsupervised depth estimation and segmentation learning.

\section{Acknowledgement}
This work was supported by D-CORE Grant from Microsoft Research Asia and partially supported by JST AIP Acceleration Research Grant Number JPMJCR20U3, and JSPS KAKENHI Grant Number JP19H01115. We would like to thank Sho Maeoki, Thomas Westfechtel, and Yang Li for the helpful discussions. We are also grateful for the GPU resources provided by Microsoft Azure Machine Learning.

{\small
\bibliographystyle{ieee_fullname}
\bibliography{egbib}
}

\appendix


\begin{center}
\textbf{\large Supplemental Materials: Neural Articulated Radiance Field}
\end{center}

\begin{figure*}
\begin{center}
\includegraphics[width=0.8\linewidth]{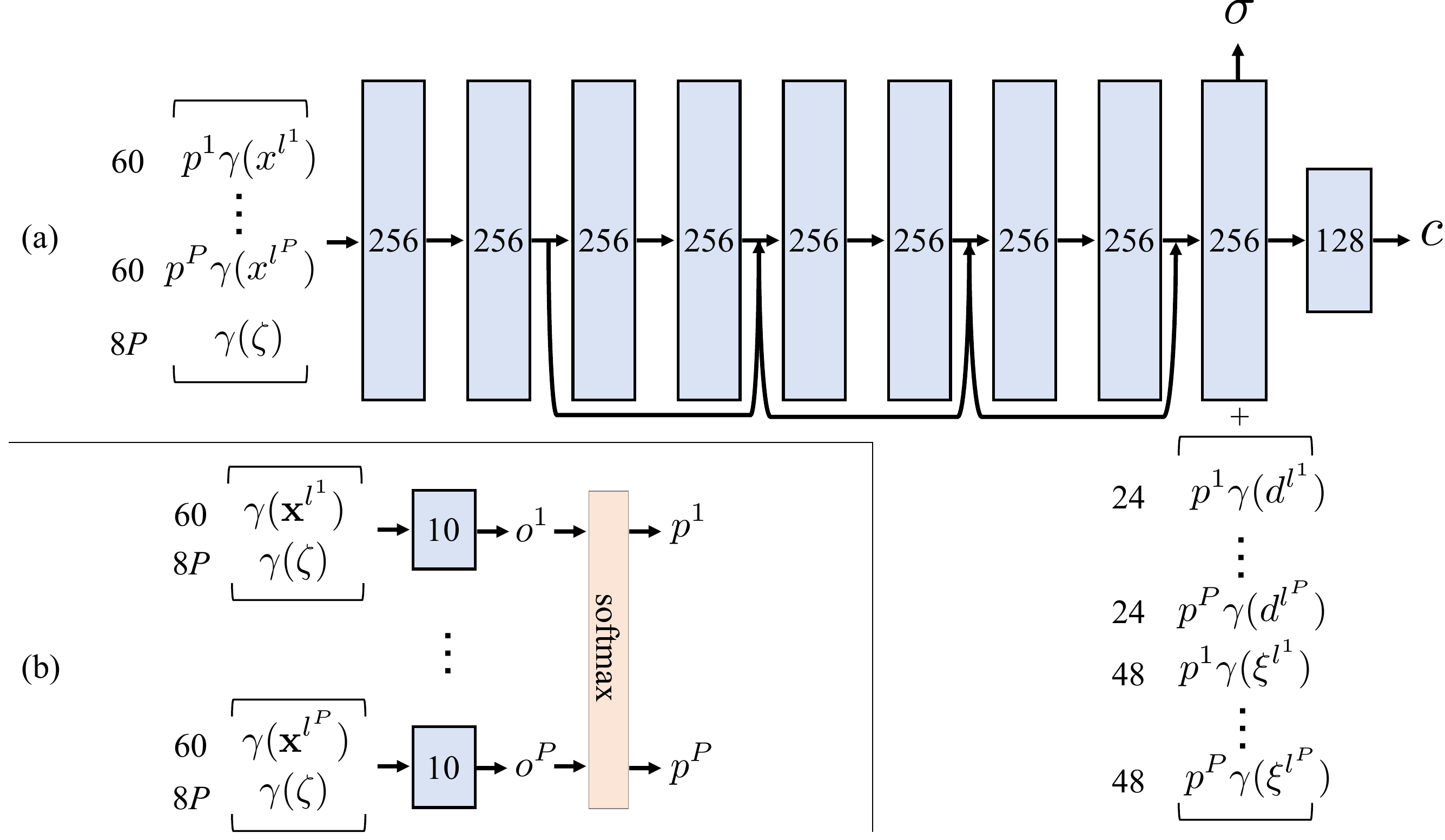}
\end{center}
\caption{Network architecture of Disentangled NARF in our experiments. (a) Network architecture of $F_{\Theta_\sigma}^{l, \zeta}$ and $F_{\Theta_c}^{l, \zeta}$ in Eqs.~22 and 23 in the main paper, modified from the original NeRF architecture diagram. `$+$' represents concatenation operation. (b) Network architecture of the selector $\mathcal{S}$.}
\label{fig:network_architecture}
\end{figure*}

\section{Ablation Studies for AutoEncoder}

\begin{figure*}
\begin{center}
\includegraphics[width=1.0\linewidth]{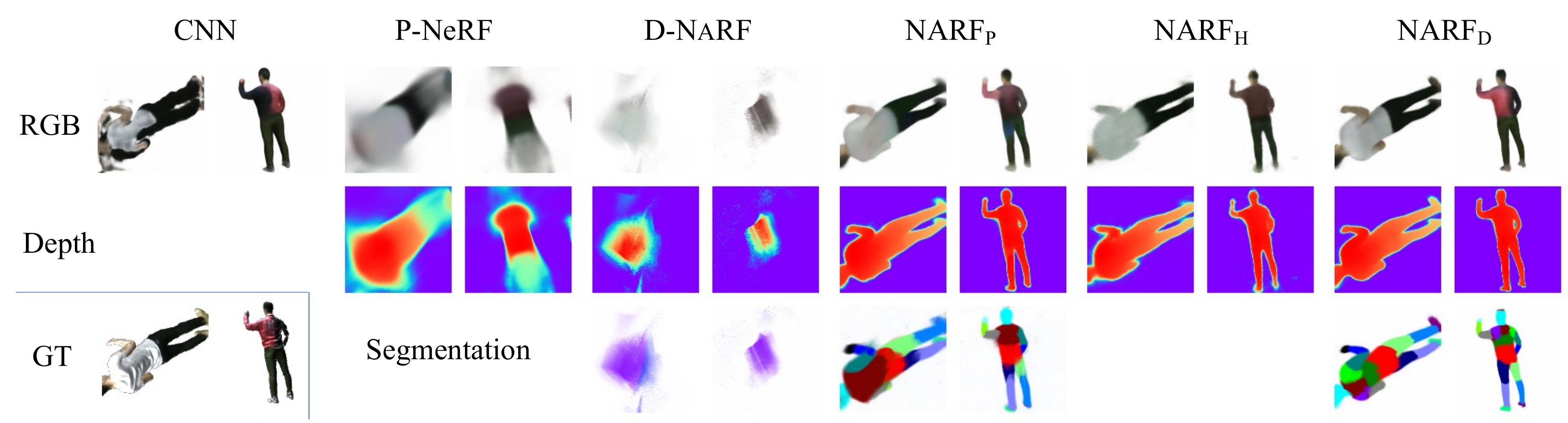}
\end{center}
\caption{Generative results comparison for AutoEncoder}
\label{fig:autoencoder_result_comp}
\end{figure*}

\begin{table*}[t]
\begin{center}
\scalebox{0.7}{
  \begin{tabular}{l|ccc|rrr|rrr|rrr|rrr} \hline
  & \multicolumn{3}{c|}{Cost} & \multicolumn{3}{c|}{Same pose, same view} & \multicolumn{3}{c|}{Novel pose, same view} & \multicolumn{3}{c|}{Same pose, novel view} & \multicolumn{3}{c}{Novel pose, novel view} \\ \hline
   Method & \#Params & \#FLPOS & \#Memory & Mask$\downarrow$ & PSNR $\uparrow$ & SSIM $\uparrow$ &  Mask $\downarrow$ & PSNR $\uparrow$ & SSIM $\uparrow$ & Mask $\downarrow$ & PSNR $\uparrow$ & SSIM $\uparrow$ & Mask $\downarrow$ & PSNR $\uparrow$ & SSIM $\uparrow$ \\ \hline \hline
CNN                           &15.6M& -   &-                            & {\bf 89.0}&{\bf 25.59}&{\bf 0.8966}&{\bf 157.2}&{\bf 24.70}&{\bf 0.8757}&385.0&{\bf 22.70}&0.8213&381.5&{\bf 22.98}&0.8243\\
P-NeRF                        &0.85M&156M &356K                         & 1526.6&19.24&0.5362&1572.1&19.60&0.5346&1911.3&18.13&0.4666&1963.2&19.08&0.4746\\
D-N{\footnotesize A}RF        &0.66M&121M &382K                         & 2067.2&18.27&0.6733&2060.2&18.81&0.6706&2026.7&18.00&0.5983&2049.5&18.82&0.6121\\  \hline
$\text{NARF}_P    $           &11.8M&\underline{2140M}&\underline{6544K}& 159.5&22.57&0.8250&182.3&22.58&0.8211&196.4&21.32&0.8056&205.0&21.80&0.8088\\
$\text{NARF}_H$               &1.06M&197M &344K                         & 201.1&22.89&0.8244&225.1&23.14&0.8205&265.1&21.48&0.7891&275.1&22.36&0.7961\\
$\text{NARF}_D$               &1.10M&205M &382K                         & 123.4&23.84&0.8568&{\bf 163.5}&23.55&0.8435&{\bf 166.2}&22.25&{\bf 0.8313}&{\bf 186.3}&22.81&{\bf 0.8294}\\
\hline
  \end{tabular}}
    \caption{Quantitative comparison for autoencoders. Best results in {\bf bold}.
    }
    \label{tb:autoencoder_result}
\end{center}
\end{table*}

In Table 1 of the main paper, we quantitatively evaluate our model in the case of a single articulated 3D object, while in Section 5 in the main paper, we only show qualitative results in the case of Autoencoder. Here, we supplement those results with the quantitative ablation study for Autoencoder in Table~\ref{tb:autoencoder_result}. The same test data settings (namely ``same pose/same view'', ``novel pose/same view'', ``same pose/novel view'', and ``novel pose/novel view''), metrics (namely PSNR, SSIM, and Mask), and baseline methods (namely $\text{NARF}_P$, $\text{NARF}_H$ and $\text{NARF}_D$, \emph{\textbf{CNN-based}}, \emph{\textbf{P-NeRF}} and \emph{\textbf{D-N{\footnotesize A}RF}}) are used for comparison. The same dataset as in Section 5 of the main paper is used for experimentation.

At testing phase, when extracting the latent shape and appearance vectors (${\bf z}_s$ and ${\bf z}_a$) using the encoder, we use images under the same viewpoint distribution as in the training images as input. Then, images from novel views and poses are rendered by combining ${\bf z}_s$ and ${\bf z}_a$ with unseen views and poses in the training data.

\paragraph{Results} Quantitative comparison results are given in Table~\ref{tb:autoencoder_result}. Qualitative results under the novel pose/novel view setting are shown in Fig.~\ref{fig:autoencoder_result_comp}. Consistent with the case of a single object, our method $\text{NARF}_D$ outperforms the others under all the evaluation metrics and test data settings (best results shown in \textbf{bold}). High quality depth and segmentation images are jointly generated as shown in Fig.~\ref{fig:autoencoder_result_comp} (rightmost column).
CNN based models cannot represent 3D structure effectively, so the performance drops significantly in the ``novel view'' settings. Fig.~\ref{fig:autoencoder_result_comp} (left-top) shows that in the novel view testing, the CNN-based method produces fuzzy images. Meanwhile, it cannot generate depth and segmentation images.
P-NeRF and D-N{\footnotesize A}RF fail in almost all settings due to implicit transformation and part dependency problems.
$\text{NARF}_P$ generally performs well, but the computational cost is too high. $\text{NARF}_H$ has poor performance on ``novel pose'' due to part dependency issues. The performance drop on novel pose in the Autoencoder case is not as significant as in the single object case (shown in Table 1 of the main paper) since the pose diversity in the training data is much larger in the Autoencoder case.

\section{Ablation Studies in RT-NeRF}
In Section 3.3 of the main paper, we introduced the rigidly
transformed neural radiance field (RT-NeRF) to effectively model a rigidly transformed object part. Here, we evaluate the effectiveness of the two most critical design elements in RT-NeRF. The first is the \emph{\textbf{explicit transformation}} that converts a global 3D location into the \emph{local} coordinate system and the local 3D location is then used to estimate the density using Eqs.~9 and 10 of the main paper. The second is the \emph{\textbf{pose-dependent color}} estimation defined in Eq.~11 of the main paper. It takes the 6D vector $\mathfrak{se}(3)$ representation $\boldsymbol \xi$ of transformation $l$ as a network input to estimate the RGB color $c$. To this end, two more baseline methods are introduced accordingly to compare to \emph{\textbf{RT-NeRF}}. The first is the rigid pose conditioned NeRF (\emph{\textbf{RP-NeRF}}) that takes the global 3D location and the rigid transformation $\boldsymbol \xi$ as network inputs, similar to the P-NeRF defined in Eq.~8 of the main paper. The second is \emph{\textbf{RT-NeRF w/o $\boldsymbol \xi$}} that estimates the RGB color $c$ without using the transformation $\boldsymbol \xi$ as input in Eq.~11 of the main paper.

\paragraph{Dataset} We create a synthetic rigid object dataset of a rendered bulldozer using Blender (a software for rendering) for experimentation.
In the dataset, the object (a bulldozer) can rigidly transform in the world coordinate system. For each rendered image, both rigid transformation and the camera viewpoint are randomly set. The camera will be translated to point to the center of the object so that the object will appear in the center of the rendered image.
The resolution for all rendered images is set to 200 $\times$ 200. In total, 480 images are used for training and another 20 images are used for testing.
The loss function is the same as in Eq.~25 of the main paper.

\begin{figure}
\begin{center}
\includegraphics[width=1.0\linewidth]{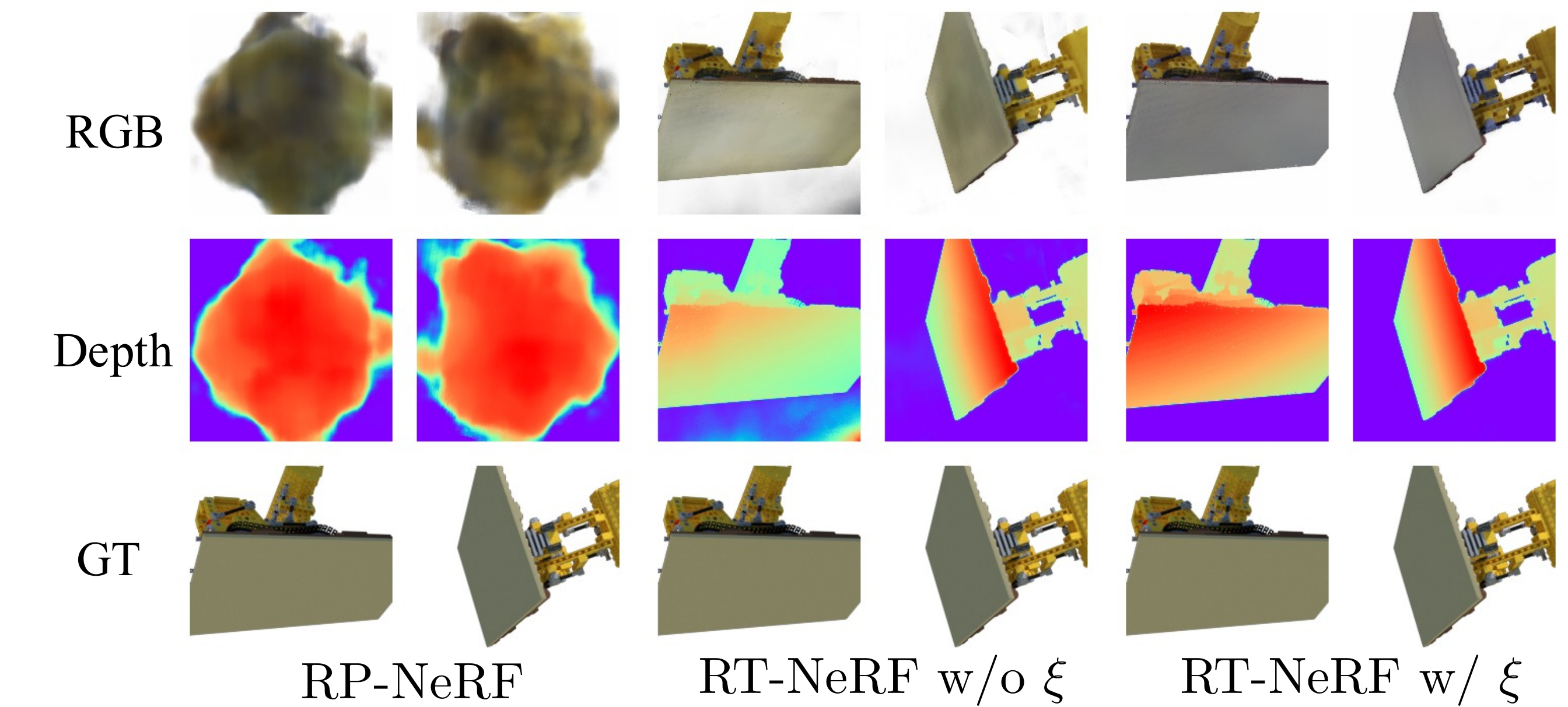}
\end{center}
\caption{Qualitative results for RT-NeRF comparison}
\label{fig:bulldozer}
\end{figure}

\paragraph{Results} The quantitative results are shown in Table~\ref{tb:rt_nerf} and the qualitative results are shown in Fig.~\ref{fig:bulldozer}.
The experimental results show that RP-NeRF is unable to learn a good 3D representation and fails to generalize to novel poses and views. In contrast, RT-NeRF effectively models the rigidly transformed object by \emph{explicitly} transforming the global 3D location into the local coordinate system.
In addition, the color estimation without the transformation input is less effective. This is concluded by comparing the results of ``RT-NeRF w/o $\xi$'' with the results of ``RT-NeRF''. Quantitatively, the performance of ``RT-NeRF w/o $\xi$'' drops significantly under the Mask metric in Table~\ref{tb:rt_nerf}. Qualitatively, the rendered images from ``RT-NeRF w/o $\xi$'' look blurry compared to ``RT-NeRF'' in Fig.~\ref{fig:bulldozer}.

\begin{table}[t]
\begin{center}
\scalebox{1.0}{
  \begin{tabular}{l|rrr} \hline
   Method & Mask$\downarrow$ & PSNR $\uparrow$ & SSIM $\uparrow$ \\ \hline \hline
RP-NeRF             & 4511.0&11.25&0.3103\\
RT-NeRF w/o $\xi$  & 29.2&19.83&0.8255\\
RT-NeRF & {\bf 16.9}&{\bf 20.05}&{\bf 0.8388}\\
\hline
  \end{tabular}}
    \caption{RT-NeRF quantitative comparison. Best results in {\bf bold}.
    }
    \label{tb:rt_nerf}
\end{center}
\end{table}

\section{Additional Ablation Studies in NARF}

\begin{table*}[t]
\begin{center}
\scalebox{0.62}{
  \begin{tabular}{|l|ccc|rrr|rrr|rrr|rrr|} \hline
  & \multicolumn{3}{c|}{Cost} & \multicolumn{3}{c|}{Same pose, same view} & \multicolumn{3}{c|}{Novel pose, same view} & \multicolumn{3}{c|}{Same pose, novel view}& \multicolumn{3}{c|}{Novel pose, novel view} \\ \hline
   Method & \#Params & \#FLPOS & \#Memory &Mask L2 $\downarrow$ & PSNR $\uparrow$ & SSIM $\uparrow$ &  Mask L2 $\downarrow$ & PSNR $\uparrow$ & SSIM $\uparrow$ & Mask L2 $\downarrow$ & PSNR $\uparrow$ & SSIM $\uparrow$  & Mask L2 $\downarrow$ & PSNR $\uparrow$ & SSIM $\uparrow$ \\ \hline \hline
CNN                           &15.6M& -   &-                &76.9&29.12&0.9429&134.8&27.30&0.9211&365.9&25.19&0.8532&392.2&24.53&0.8470\\
P-NeRF                        &0.85M&156M &356K             &778.7&21.42&0.8006&1077.0&20.42&0.7696&844.9&21.19&0.7897&1110.1&20.27&0.7648\\
D-N{\footnotesize A}RF        &0.66M&121M &382K             &2182.6&18.90&0.1143&2308.2&18.81&0.1140&2137.3&19.09&0.1144&2241.3&18.88&0.1133\\ \hline
$\text{NARF}_P$   &11.8M&2149M&6544K&92.0&28.56&0.9258&116.2&26.83&0.9052&101.5&27.54&0.9144&125.8&26.50&0.9104\\
$\text{NARF}_H$               &1.06M&197M &344K             &55.6&29.91&0.9470&376.8&24.09&0.8665&70.5&28.81&0.9370&374.6&23.98&0.8646\\
$\text{NARF}_D$               &1.10M&205M &382K             &{\bf 50.5}&{\bf 30.86}&{\bf 0.9586}&{\bf 114.4}&{\bf 27.93}&{\bf 0.9317}&{\bf 64.1}&{\bf 29.44}&{\bf 0.9466}&{\bf 123.8}&{\bf 27.24}&{\bf 0.9230}\\ \hline
$\text{NARF}_D sigmoid$       &1.10M&205M &382K             & 50.8&30.74&0.9578&117.7&27.83&0.9304&64.6&29.35&0.9459&125.4&26.17&0.9129\\     \hline
$\text{NARF}_P 32$            &0.32M&59M  &824K             & 114.8&27.85&0.9191&139.9&26.86&0.9068&122.9&27.14&0.9105&147.3&25.74&0.8933\\
$\text{NARF}_P 64$            &0.98M&178M &1641K            & 105.5&27.86&0.9212&126.9&26.78&0.9094&113.8&27.21&0.9127&133.1&26.28&0.9041\\
$\text{NARF}_P 128$           &3.28M&596M &3275K            & 93.8&28.46&0.9284&116.1&27.15&0.9135&102.9&27.56&0.9180&124.2&26.23&0.9068\\
$\text{NARF}_P 256$           &11.8M&2149M&6544K            & 92.0&28.56&0.9258&116.2&26.83&0.9052&101.5&27.54&0.9144&125.8&26.50&0.9104\\     \hline
$\text{NARF}_P, \tau=0.01$    &0.32M&59M  &824K             & 240.1&25.81&0.8730&282.0&24.92&0.8541&256.0&25.32&0.8624&287.9&23.72&0.8333\\
$\text{NARF}_P, \tau=0.1$     &0.32M&59M  &824K             & 168.1&26.79&0.8995&196.7&25.92&0.8863&176.1&26.26&0.8911&204.8&25.10&0.8756\\
$\text{NARF}_P, \tau=1$       &0.32M&59M  &824K             & 141.5&27.24&0.9077&169.9&26.29&0.8949&150.9&26.56&0.8989&178.2&25.16&0.8809\\
$\text{NARF}_P, \tau=10$      &0.32M&59M  &824K             & 114.8&27.85&0.9191&139.9&26.86&0.9068&122.9&27.14&0.9105&147.3&25.74&0.8933\\
$\text{NARF}_P, \tau=100$     &0.32M&59M  &824K             & 107.9&27.83&0.9190&130.2&26.96&0.9059&115.9&27.23&0.9117&137.5&26.24&0.8983\\
$\text{NARF}_P, \tau=1000$    &0.32M&59M  &824K             & 109.0&27.44&0.9137&130.8&26.68&0.9014&117.0&26.91&0.9072&138.3&26.08&0.8953\\    \hline
$\text{NARF}_P$ w/o $L_{mask}$&0.32M&59M  &824K             & 193.9&28.01&0.9131&220.2&26.72&0.8994&208.7&27.33&0.9028&229.8&25.54&0.8792\\
$\text{NARF}_H$ w/o $L_{mask}$&1.06M&197M &344K             & 6816.2&24.25&0.7004&6893.1&21.10&0.6636&6876.3&23.40&0.7029&6957.2&20.53&0.5244\\
$\text{NARF}_D$ w/o $L_{mask}$&1.10M&205M &382K             & 113.4&{\bf 31.39}&{\bf 0.9630}&165.7&{\bf 28.09}&{\bf 0.9346}&132.8&{\bf 29.90}&{\bf 0.9508}&263.7&25.71&0.8889\\   \hline
  \end{tabular}}
    \caption{Quantitative results of ablation studies.}
    \label{ablation}
\end{center}
\end{table*}

In this section, we provide more ablation studies on the effects of the mask loss (the second item in Eq.~25 of the main paper), temperature parameter for $\text{NARF}_P$ (in Eq.~15 of the main paper), and softmax activation function for $\text{NARF}_D$ (in Eq.~21 of the main paper).

\begin{figure*}
\begin{center}
\includegraphics[width=1.0\linewidth]{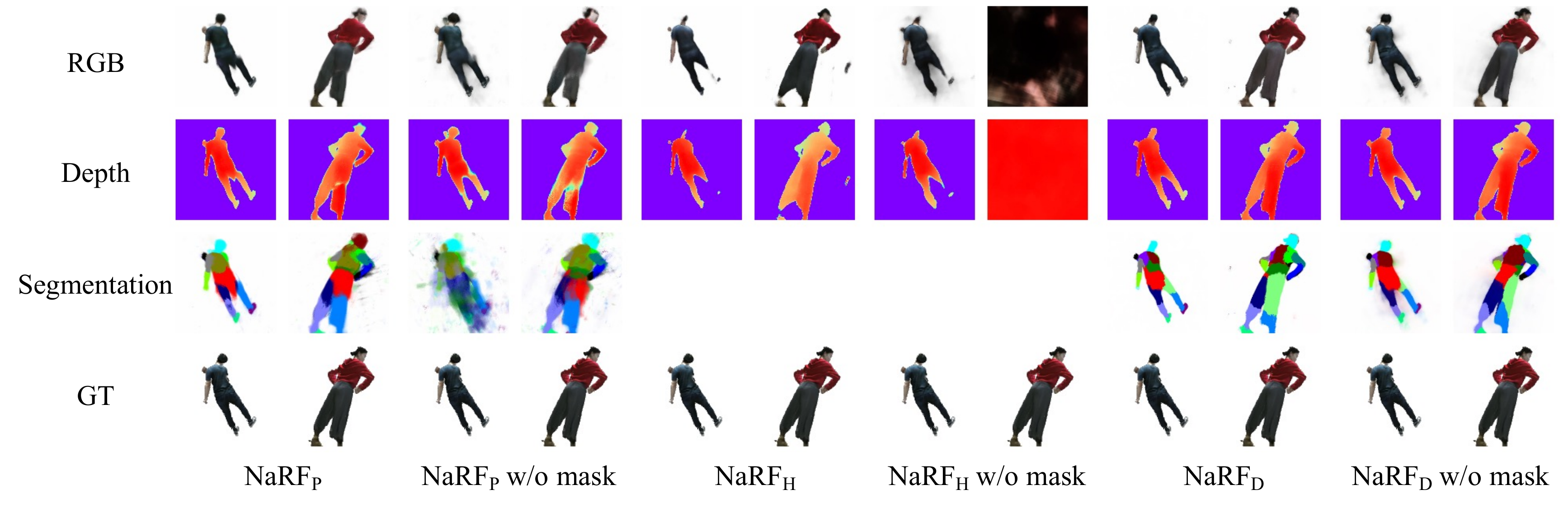}
\end{center}
\caption{Comparison of mask loss.}
\label{fig:mask_comparison}
\end{figure*}

\paragraph{W/ and w/o mask loss}
We conduct experiments to evaluate the effect of the mask loss (the second term of Eq.~25 of the main paper) added to the rendered mask image. While the color loss optimizes the final RGB color of the rendered pixels, the gradient from the mask loss directly optimizes the densities of the 3D locations on the camera rays. Therefore, the additional mask loss is helpful in learning 3D shapes efficiently. The quantitative and qualitative results of w/ and w/o mask loss are shown in Table~\ref{ablation} and Fig.~\ref{fig:mask_comparison} respectively. In Table~\ref{ablation}, it can be seen that performances of all the three variants of NARF drop significantly without the mask loss, especially under the novel pose/novel view setting. Particularly, in Fig.~\ref{fig:mask_comparison}, the $\text{NARF}_H$ model is not able to converge at all without the mask loss. The $\text{NARF}_P$ and $\text{NARF}_D$ models still work without the mask loss but the rendered images get very blurry, especially on the background regions around the object.

\begin{figure*}
\begin{center}
\includegraphics[width=1.0\linewidth]{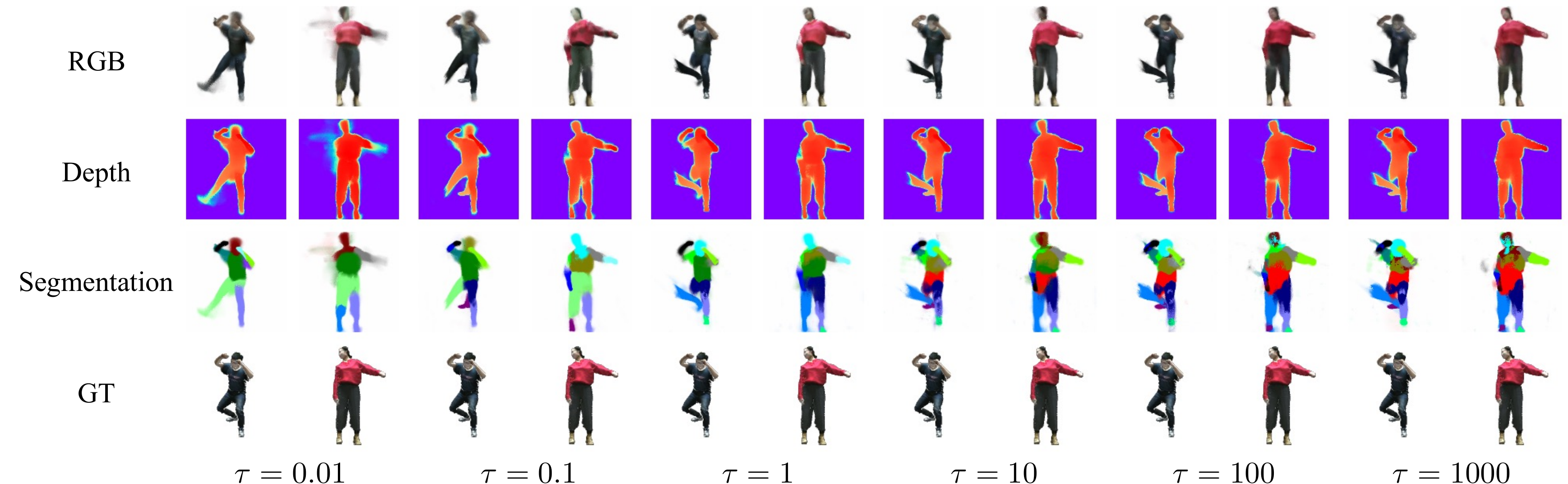}
\end{center}
\caption{Comparison of temperature $\tau$ of $\text{NARF}_P$.}
\label{fig:tmp_comparison}
\end{figure*}

\paragraph{Temperature parameter $\tau$ in $\text{NARF}_P$}
We study the effect of the temperature parameter $\tau \in (0, \infty)$ in $\text{NARF}_P$ (in Eq.~15 of the main paper). The temperature parameter determines how soft the selection is among the multiple RT-NeRFs. When $\tau$ is close to $0$, hard selection is performed. Though the \emph{\textbf{Part Dependency}} prior is strictly satisfied in this case, convergence in the training is difficult since the highest current estimate will completely block the gradient from back-propagating to the others. It is especially worse in the early stage of the training when the highest estimate is almost random. In turn, when $\tau$ is close to $\infty$, averaging is performed. In this case, the gradient is back-propagated to all RT-NeRFs, but the \emph{\textbf{Part Dependency}} issue arises again, which will harm the generalization ability to novel poses.
The quantitative and qualitative results are shown in Table~\ref{ablation} and Fig.~\ref{fig:tmp_comparison}, respectively. We empirically use the best-performing $\tau=100$ setting as shown in Table~\ref{ablation}.

\paragraph{Softmax vs. sigmoid activation in $\text{NARF}_D$}
In Eq.~21 of the main paper, we use softmax activation, which is also motivated by the \emph{\textbf{Part Dependency}} prior. Here, we provide the results of using the sigmoid activation as an alternative. Formally, Eq.~21 of the main paper is replaced with Eq.~\ref{eq:sigmoid_selector}:
\begin{align}\label{eq:sigmoid_selector}
    &O_{\Gamma}^i: (\gamma({\bf x}^{l^i}), \gamma(\boldsymbol \zeta)) \rightarrow (o^i), \,
    &p^i = \frac{1}{1 + \exp(-o^i)}
\end{align}
The quantitative results are shown in Table~\ref{ablation}.
In conclusion, softmax outperforms sigmoid, especially in the ``novel pose/novel view'' setting.

\section{Implementation Details in P-NeRF}
In Eq.~8 of the main paper, P-NeRF takes a global 3D location ${\bf x}$ and the part transformations $\{l^i|i=1,...,P\}$ as input. For implementation, we use the 6D vector $\mathfrak{se}(3)$ representation ${\boldsymbol \xi}^i$ of transformation $l^i$, and concatenate them together with ${\bf x}$ and the bone length $\zeta$ as the network input. Positional encoding is performed before the concatenation. Formally,
\begin{align}
&F_{\Theta}^{\mathcal{P}}: (\gamma({\bf x}), \{\gamma(\xi^i)|i=1,...,P\}, \gamma(\zeta), \gamma({\bf d})) \rightarrow (\sigma, c).
\label{eq:p-nerf}
\end{align}
The density and color sub-networks are defined as
\begin{align}
&F_{\Theta_\sigma}^{\mathcal{P}}: (\gamma({\bf x}), \{\gamma(\xi^i)|i=1,...,P\}, \gamma(\zeta)) \rightarrow (\sigma, {\bf h}),\\
&F_{\Theta_c}^{\mathcal{P}}: ({\bf h}, \{\gamma(\xi^i)|i=1,...,P\}, \gamma({\bf d})) \rightarrow (c).
\label{eq:p-nerf2}
\end{align}

\section{Implementation Details in D-N{\footnotesize A}RF}

D-N{\footnotesize E}RF~\cite{nerfies} uses a canonical template and learns the observation-to-canonical deformation
\begin{equation}
    \Psi: ({\bf x}, \omega) \rightarrow {\bf x}'
\end{equation}
where $\omega$ is a deformation latent code.

D-N{\footnotesize E}RF is defined on the deformed position ${\bf x}'$ in the canonical template,
\begin{equation}
    G: (\Psi({\bf x}, \omega), {\bf d}, \psi)\rightarrow(\sigma, c)
\end{equation}
where $\psi$ is a latent appearance code.

In our case, $\omega$ and $\psi$ correspond to the pose configuration $\mathcal P$ and the appearance latent vector ${\bf z}_a$ respectively.
$\Psi$ is implemented using MLP, which seems to suffer from the problem of implicit transformation. In our setup, the pose of each part is given, so it can be implemented more directly.
In our implementation, we first use the occupancy network similar to the one defined in Eq.~21 of the main paper to decide which part the input point belongs to.
\begin{align}
    &O_{\Gamma}^i: (\gamma({\bf x}^{l^i}), \gamma(\boldsymbol \zeta)) \rightarrow (o^i), \\
    &p^i = \frac{\exp(o^i)}{\sum_{k=1}^P \exp(o^k)},
\end{align}
Then, we calculate the coordinates ${\bf x}'$ on the canonical shape as
\begin{equation}
    {\bf x}' = \sum_i ({\bf x}^{l^i} + {\bf t}_{\text{canonical}}^i) * p_i
\end{equation}
where ${\bf t}_{\text{canonical}}^i$ is the origin of a canonical pose's $i^\text{th}$ part in the global coordinate system. View direction in the canonical space and the transformation vector are defined as
\begin{align}
{\bf d} = \sum_i {\bf d}^{l^i} * p_i,\,
\boldsymbol \xi = \sum_i \boldsymbol \xi^i * p_i
\end{align}
Then, the D-N{\footnotesize A}RF we have implemented in the experiment is defined as
\begin{equation}
F_{\Theta_{\sigma}}^{l, \boldsymbol \zeta}: (\gamma({\bf x}'), \gamma(\boldsymbol \zeta)) \rightarrow (\sigma, {\bf h}),
\end{equation}
\begin{equation}
F_{\Theta_{c}}^{l, \boldsymbol \zeta}: ({\bf h}, \gamma({\bf d}), \gamma(\boldsymbol \xi)) \rightarrow (c).
\end{equation}

This implementation is similar to the implementation of $\text{NARF}_D$, differing only in how the coordinates are input to the model. The results of the experiments show that a concatenation-based $\text{NARF}_D$ model that retains the coordinates for all parts is more effective than a transformation on the input coordinates in D-N{\footnotesize A}RF.

\section{Training Details}
We used the Adam \cite{kingma2014adam} optimizer with an initial equalized learning rate \cite{karras2018progressive} of $0.01$.
The learning rate is decayed to $0.99995\times$ of the previous iteration. Particularly, P-NeRF and $\text{NARF}_H$ based autoencoders are trained with an initial learning rate of 0.001 since the training will explode if a learning rate of 0.01 is used. The batch size is set to 16 for all experiments. We sample as many camera rays as can be fit in the GPU memory. The training converges at about 100,000 iterations. The training of our method $\text{NARF}_D$ takes 24 hours on 4 V100 GPUs. The code for creating our synthetic datasets is available at \textcolor[rgb]{1,0,1}{https://github.com/nogu-atsu/NARF}.

\section{Cross Dataset Evaluation on SURREAL Dataset}
\begin{figure}[t]
\begin{center}
\includegraphics[width=1.0\linewidth]{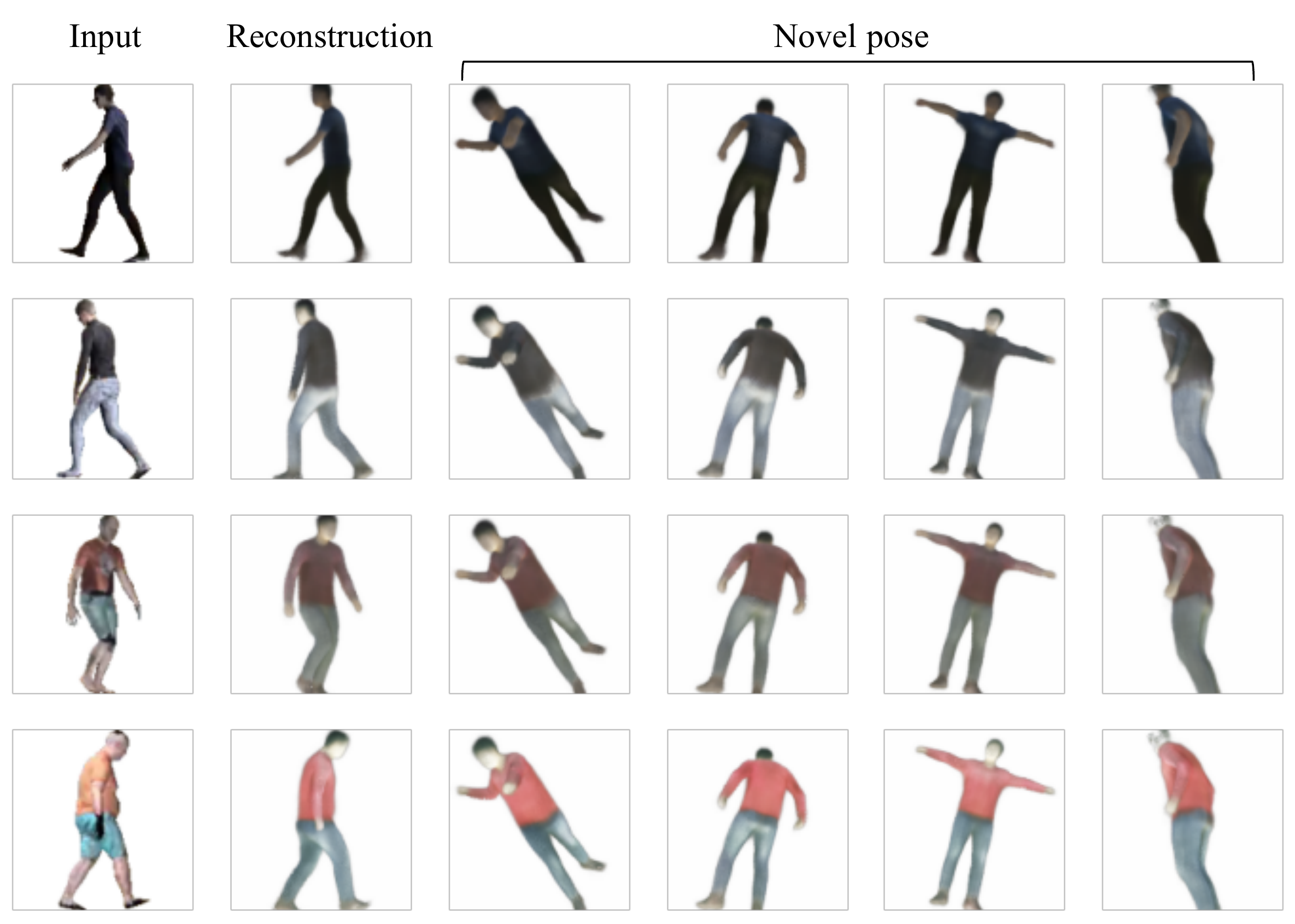}
\end{center}
\caption{Qualitative results on SURREAL dataset. The cross dataset protocol is used for evaluation.}
\label{fig:surreal_results}
\end{figure}

In order to verify the generalization ability of NARF autoencoder across different datasets, we use a cross dataset evaluation protocol that trains the model on THUman dataset~\cite{Zheng2019DeepHuman}, then tests it on SURREAL dataset~\cite{varol17_surreal}. There are several differences between the THUman and SURREAL datasets.
First, even though the human samples in SURREAL contain rendered SMPL meshes and textures as in THUman, but unlike in THUman, the meshes do not contain clothing. Second, the camera and shape parameters in SURREAL, including the distance between camera and human, human pose distribution and the range of body size are quite different from that in THUman.
Even so, we test images from the test set of SURREAL using the NARF autoencoder trained on THUman.
The qualitative reconstruction results are shown in Fig.~\ref{fig:surreal_results}. From left to right, we show the input SURREAL images, their reconstructed images, and the images re-rendered under different pose configurations.
Since the THUman dataset is not diverse enough in terms of clothing and body size, the reconstructions of short pants and fat people are less effective. But the novel view and pose rendering results look quite reasonable.

\begin{figure*}[t]
\begin{center}
\includegraphics[width=1.0\linewidth]{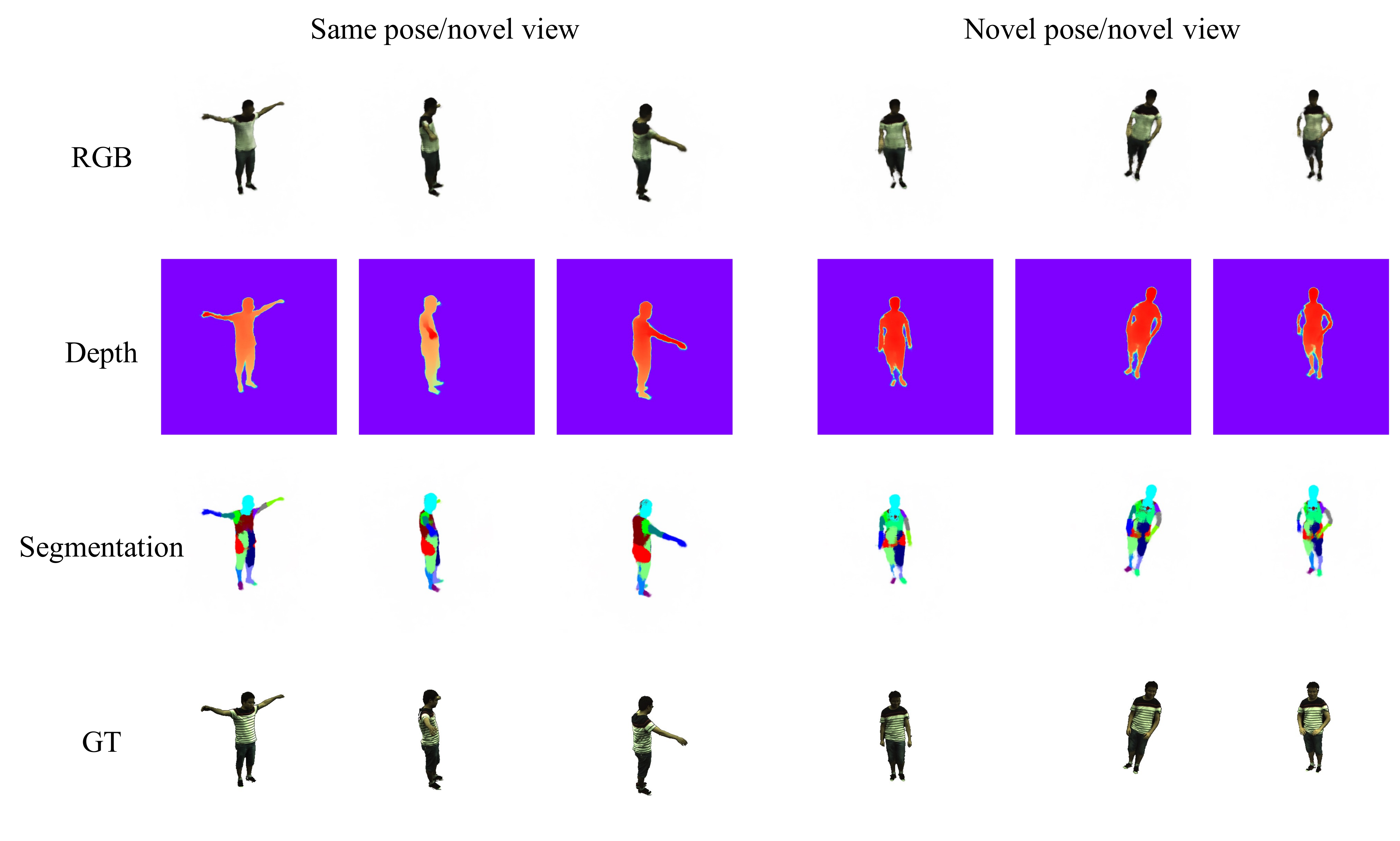}
\end{center}
\caption{Results of $\text{NARF}_D$ on real human images}
\label{fig:zju_results}
\end{figure*}

\section{Experiment on Real Human Images}
In this section, we test our approach on a real human dataset ZJU-MOCAP~\cite{peng2021neural}. ZJU-MOCAP
is a multi-view person video dataset. For each frame, SMPL parameters~\cite{loper2015smpl} are given. We use the first 1969 frames (90\%, 2185 frames in total) of the Taichi class video for training and the remaining 216 frames (10\%) for testing (novel pose). The resolution of the image is 512 $\times$ 512.

The qualitative results of $\text{NARF}_D$ on this dataset are shown in Fig.~\ref{fig:zju_results}.
The left part of Fig.~\ref{fig:zju_results} shows the pose used in the training, but rendered from novel viewpoints, and the right part of Fig.~\ref{fig:zju_results} shows the novel pose/novel view testing results.
The quality of the rendered images is not as good as testing on our synthetic datasets.
This might be caused by the assumption that the parts are rigid objects, which may not be perfectly satisfied for real images. For example, loose clothes may move when a person makes a movement.
This issue can be considered in future work, for example, by learning latent variables to account for both pose-dependent and pose-independent deformations similar to Neural Body~\cite{peng2021neural}.

Although the quality of the rendered images for a real person from our method still has room for improvement, we believe that the proposed explicitly controllable representation of viewpoint, pose, bone parameters, and appearance for the articulated object is an important contribution.

\end{document}